\documentclass[lettersize,journal]{IEEEtran}
\usepackage{amsmath,amssymb,amsfonts}
\usepackage{algorithm}
\usepackage{array}
\usepackage[caption=false,font=normalsize,labelfont=sf,textfont=sf]{subfig}
\usepackage{textcomp}
\usepackage{stfloats}
\usepackage{url}
\usepackage{verbatim}
\usepackage{graphicx}
\usepackage{cite}
\usepackage{xcolor}
\usepackage{xcolor}
\usepackage{multirow}
\definecolor{darkgreen}{rgb}{0.0, 0.5, 0.0}
\usepackage{array}
\usepackage{verbatim}

\usepackage{times}
\usepackage{bbding}
\usepackage{algpseudocode}
\algrenewcommand\alglinenumber[1]{\footnotesize #1}
\usepackage{lettrine}

\usepackage{booktabs}
\usepackage{pifont}
\usepackage{makecell}

\usepackage{setspace}

\usepackage{multirow}

\begin{document}

\title{Extending Reconstruction: Reconstruction-to-Vector Diffusion for Hyperspectral Anomaly Detection}

\author{Jijun~Xiang,
        Tao~Wang,
        Jiayi~Wang,
        Pengxiang~Wang,
        Cheng~Chen,
        and~Nian~Wang
\thanks{Jijun Xiang, Tao Wang, Jiayi Wang, Pengxiang Wang, Cheng Chen and Nian Wang are with the Rocket Force University of Engineering, Xi'an 710025, China (e-mail: jijunxiang0614@163.com; twang305@126.com; wjjy0111@163.com; 13786633097@163.com; chengchen0626@126.com; nianwang04@outlook.com).}
\thanks{Corresponding author: Tao Wang.}
}

\markboth{Journal of \LaTeX\ Class Files,~Vol.~14, No.~8, August~2021}%
{Shell \MakeLowercase{\textit{et al.}}: A Sample Article Using IEEEtran.cls for IEEE Journals}

\IEEEpubid{0000--0000/00\$00.00~\copyright~2021 IEEE}

\maketitle

\begin{abstract}
While Hyperspectral Anomaly Detection (HAD) excels at identifying sparse targets in complex scenes, conventional reconstruction-based models typically rely on a scalar metric as the final detection criterion. This heavy reliance on magnitude-only scalar residuals frequently triggers sub-pixel anomaly vanishing during spatial downsampling, alongside confirmation bias when unpurified targets corrupt background representations. In this paper, we propose Reconstruction-to-Vector Diffusion (R2VD), which extends the reconstruction paradigm by repurposing it for background manifold purification, followed by high-dimensional generative score modeling. Our framework introduces a four-stage pipeline: (1) a Physical Prior Extraction (PPE) stage that mitigates early confirmation bias via dual-stream statistical guidance; (2) a Guided Manifold Purification (GMP) stage utilizing an OmniContext Autoencoder (OCA) to extract purified residual maps while preserving fragile sub-pixel topologies; (3) a Residual Score Modeling (RSM) stage where a Diffusion Transformer (DiT), guarded by a Physical Spectral Firewall (PSF), effectively isolates cross-spectral leakage; and (4) a Vector Dynamics Inference (VDI) stage that robustly decouples targets from backgrounds by evaluating high-dimensional vector interference patterns instead of conventional scalar errors. Comprehensive evaluations on eight datasets confirm that R2VD establishes a new state-of-the-art, delivering exceptional target detectability and background suppression. The code is
available at https://github.com/Bondojijun/R2VD.
\end{abstract}


\section{Introduction}\label{1}
\IEEEPARstart{H}{yperspectral} Anomaly Detection (HAD) \cite{liu2025adaptive}, \cite{li2025dpmn}, \cite{he2025cwimamba} holds significant application value across various fields, primarily due to its unsupervised nature that requires no prior knowledge. In hyperspectral imagery, an anomaly is conventionally defined as a spatially sparse pixel or compact region whose spectral signature differs significantly from that of its surrounding background, while its material identity is not known a priori. However, in real-world Earth observation, the inherent high-dimensional heterogeneity of Hyperspectral Images (HSIs) \cite{fu2025mmr}, \cite{dwsdiff}, \cite{li2023you}, combined with non-stationary illumination, cloud occlusion, and complex local topography, makes it highly challenging to accurately decouple spatially sparse anomalies from strong background clutter.

\begin{figure}[t]
    \centering
    \includegraphics[width=0.46\textwidth]{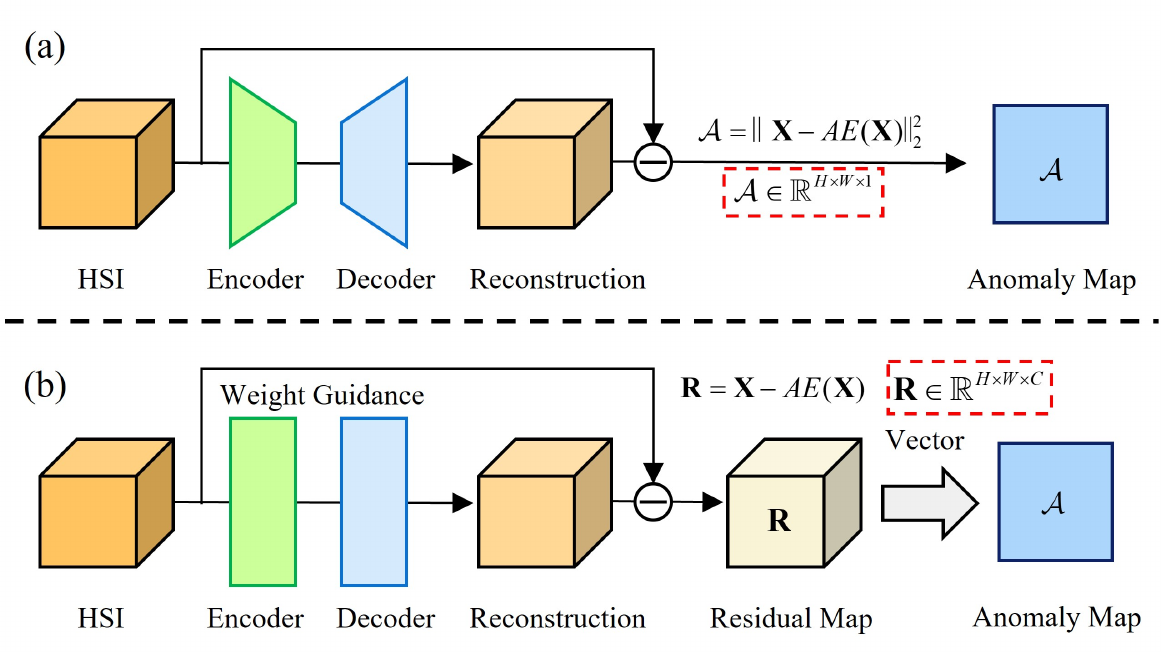}
    \caption{Comparison of HAD residual evaluation approaches. (a) Traditional scalar reconstruction using the $L_2$ norm of the residual as the detection metric. (b) Proposed Reconstruction-to-Vector pipeline, which repurposes reconstruction for manifold purification to enable subsequent high-dimensional vector dynamics inference.}
    \label{fig:reconstruction}
    \vspace*{-0.8\baselineskip}
\end{figure}

To overcome the linear representation bottlenecks of traditional statistical models, deep learning has been widely adopted in HAD. Most state-of-the-art methods follow a reconstruction paradigm based on Autoencoders (AEs) \cite{gaed}, \cite{rsaae} or Generative Adversarial Networks (GANs) \cite{sslgan}, as illustrated in Fig. \ref{fig:reconstruction}. The core logic involves unsupervised training on HSI data to capture the global background manifold, subsequently reconstructing an idealized background-only image. Anomaly scores are then determined by calculating the scalar reconstruction error (i.e., the $L_2$ norm of the residual) between the original and reconstructed images.
\IEEEpubidadjcol

Research surrounding this paradigm has evolved along three main trajectories. In terms of topological architecture, diverse frameworks have been explored, ranging from Auto-AD \cite{wang2021auto} with local receptive fields to Transformer-based GT-HAD \cite{lian2024gt} and graph-based MGAE \cite{MGAE} for global dependencies, as well as State Space Models (SSMs) based CWIMamba \cite{he2025cwimamba} and diffusion-based DWFDiff \cite{dwsdiff}. Regarding optimization mechanisms, strategies to alleviate identity mapping include the adversarial game in SSLGAN \cite{sslgan}, adaptive residual weighting in LRR \cite{LRR}, and masked self-supervised penalties in AETNet \cite{li2023you}. Furthermore, for manifold purification and prior guidance, physical constraints and denoising processes have been introduced in models such as GAED \cite{gaed}, MAAE \cite{maae}, RGAE \cite{rgae}, and BSDM \cite{ma2025bsdm}.

\begin{figure}[t]
    \centering
    \includegraphics[width=0.46\textwidth]{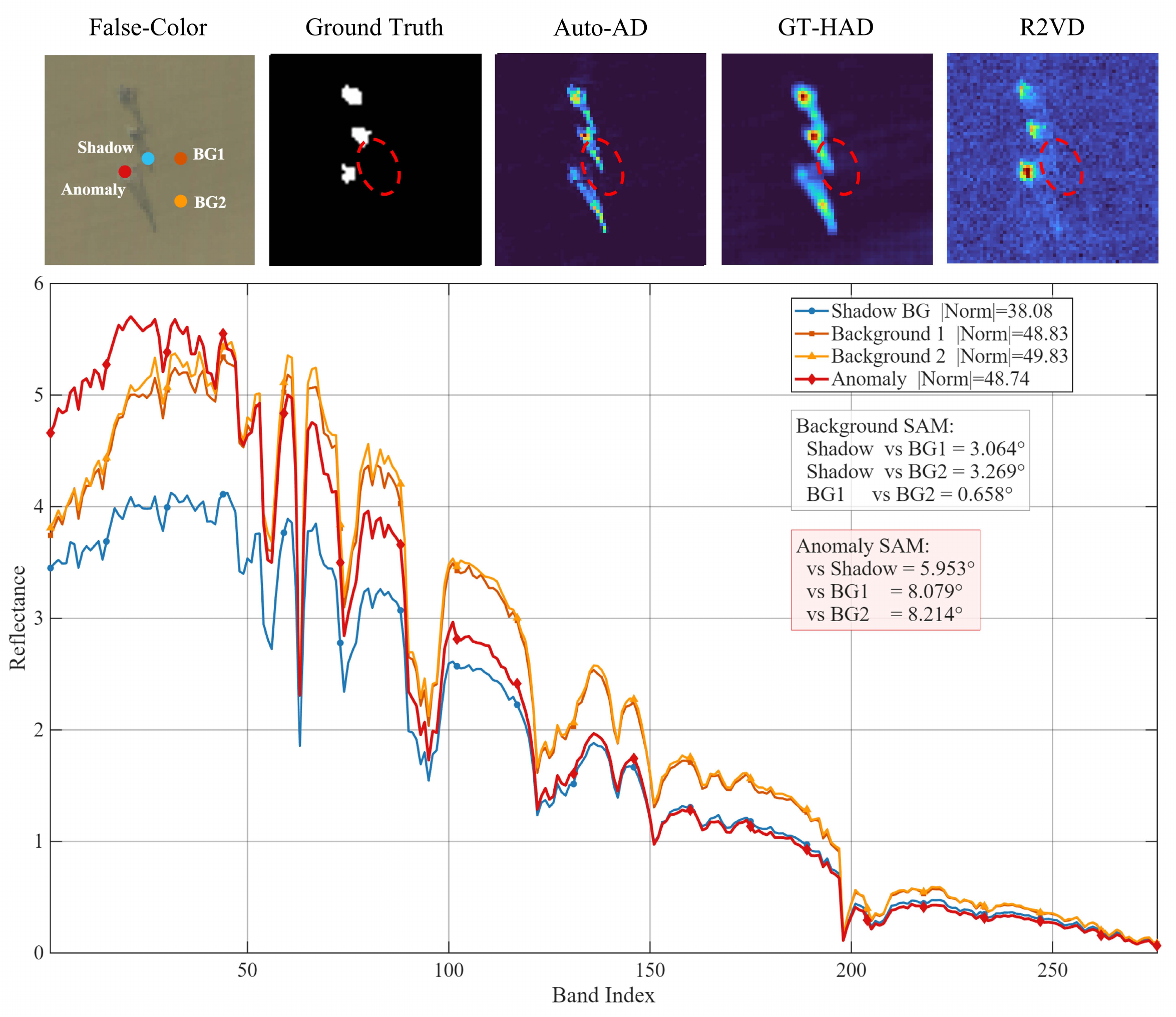}
    \vspace*{-0.5\baselineskip}
    \caption{Visual and spectral analysis on the HAD100-40 dataset. Top: Detection results comparing traditional methods and R2VD. Bottom: Spectral curves of selected pixels.}
    \vspace*{-0.9\baselineskip}
    \label{fig:shadow}
\end{figure}

Despite significant progress in background representation and feature decoupling, these evolutionary paths invariably converge on a common logical endpoint: relying on the scalar reconstruction error ($L_2$ norm) as the final detection metric. While computationally efficient, this conventional metric implicitly evaluates only the magnitude of the residual vector, thereby failing to fully exploit the rich high-dimensional directional and structural information embedded in HSIs. Consequently, scalar residuals exhibit inherent ambiguity when distinguishing genuine high-contrast anomalies from complex shadow false alarms induced by illumination variations. As shown in Fig. \ref{fig:shadow} on the HAD100-40 dataset, both Auto-AD and GT-HAD erroneously misclassify shadow regions as anomalies. Spectral analysis reveals that although the shadow region's magnitude (Norm = 38.08) is significantly lower than that of the normal background (Norm $\approx$ 49), the spectral angle between the shadow and the background (SAM $\approx$ 3$^\circ$) is substantially smaller than that between the anomaly and the background (SAM $\approx$ 8$^\circ$). This confirms that shadows and the background share striking directional consistency in the manifold space. By relying solely on magnitude energy differences while overlooking such angular consistency, standard scalar metrics remain susceptible to these local illumination-induced false alarms. 

Furthermore, implementing the traditional reconstruction-based paradigm in complex real-world scenes encounters two practical bottlenecks. First, models frequently employ spatial downsampling to improve computational efficiency and expand the receptive field. However, the low-pass filtering effect induced by this downsampling inadvertently smooths out high-frequency sub-pixel anomaly information, causing fragile anomalies to lose their discriminative features and vanish in the residual space. Second, unsupervised training without explicit prior guidance is highly susceptible to confirmation bias. In this work, confirmation bias refers to an unsupervised learning pathology where the network, driven by unconstrained reconstruction error minimization, erroneously encodes and memorizes the features of anomalous targets, implicitly treating them as background context. Driven by the unconstrained minimization of reconstruction errors, the network tends to encode and memorize the features of anomalous targets, implicitly treating them as background context.

To alleviate these limitations, we propose the Reconstruction-to-Vector Diffusion (R2VD) framework. Instead of relying on scalar residuals as the final metric, R2VD decouples the pipeline into consecutive stages: utilizing reconstruction solely for background manifold purification, followed by a generative model for high-dimensional vector dynamics inference. Specifically, to resolve sub-pixel attenuation, we introduce Residual Score Modeling (RSM) and Vector Dynamics Inference (VDI) based on a full-resolution Diffusion Transformer (DiT) \cite{peebles2023scalable}. Constrained by a Physical Spectral Firewall (PSF), the DiT models the score field directly in the purified residual space to suppress cross-spectral leakage. Instead of relying on scalar thresholding, VDI evaluates vector consistency under $K$ random perturbations \cite{ho2020denoising}, utilizing score vector stability under random dispersion as a discriminative metric. For the confirmation bias loop, the Guided Manifold Purification (GMP) stage employs an OmniContext Autoencoder (OCA) to enlarge the receptive field without sacrificing spatial resolution, preserving sub-pixel topologies. GMP is initialized by a Physical Prior Extraction (PPE) stage, which aligns coarse anomaly scores from classical RX \cite{RX} and LSUN \cite{LSUN} algorithms via Percentile Rank Alignment (PRA). This initial physical weight map is dynamically self-corrected within the autoencoder, preventing the network from memorizing anomalies. Consequently, R2VD extends the reconstruction paradigm by shifting the final decision from ambiguous scalar errors to refined vector dynamics.

Our main contributions are summarized as follows:
\begin{itemize}
    \item We propose the R2VD framework, which extends the reconstruction-based HAD paradigm by shifting the final evaluation from ambiguous scalar residuals to refined score-field vector dynamics to exploit high-dimensional directional consistency.
    \item We design a progressive boundary tightening loop governed by an asymmetric uncertainty gating mechanism. Initialized by complementary statistical and subspace priors , it explicitly breaks the confirmation bias loop to yield an uncontaminated background manifold.
    \item We introduce an OCA paired with a PSF-constrained DiT. This architecture eliminates spatial downsampling to preserve fragile sub-pixel topologies while preventing cross-spectral feature leakage.
\end{itemize}

\section{Related Work}\label{sec:rw}

\subsection{Deep Learning-based Hyperspectral Anomaly Detection} 

Traditionally, HAD relies on statistical models, such as RX \cite{RX}, LRX \cite{LRX}, and KRX \cite{KRX}, alongside representation-based techniques like CRD \cite{CRD} and LRR \cite{LRR}, and subspace methods including OSP-TD \cite{OSP-TD} and LSUN \cite{LSUN}. However, these classical paradigms are fundamentally constrained by rigid statistical assumptions that falter in complex scenes, coupled with the inadequate feature abstraction capabilities of handcrafted linear models. To overcome these bottlenecks, deep learning drives a continuous paradigm shift toward robust representation frameworks. Early approaches establish reconstruction-based baselines using Autoencoders (AEs) and Convolutional Neural Networks (CNNs), with models like Auto-AD \cite{wang2021auto} and DFAN \cite{cheng2024deep} focusing on extracting local neighborhood priors. To further model complex background manifolds, Generative Adversarial Networks (GANs) leverage adversarial training to improve discriminative reconstruction. Representative methods include GANHAD \cite{GANHAD}, the frequency-to-spectrum mapping GAN (FTSGAN) \cite{FTSGAN}, and CL-CaGAN \cite{CL-CaGAN}. In particular, FTSGAN reformulates background reconstruction as a frequency-to-spectrum mapping task, where spectra are mapped from the fractional Fourier domain back to the original spectral domain. In parallel, self-supervised contextual reconstruction methods have been developed to alleviate anomaly fitting by preventing the network from directly observing the pixel or region being reconstructed. This branch includes blind-spot and blind-block architectures, such as BockNet \cite{BockNet}, DirectNet \cite{SDWIRN}, and the blind-spot KAN-based background reconstruction network (BKP-Net) \cite{BSKBRN}. By masking potentially anomalous local content during reconstruction, these methods encourage the reconstructed output to conform to the background distribution and consequently enlarge the residual response of anomalous targets. As the demand for capturing global spatial-spectral dependencies grows, convolutional constraints are superseded by attention mechanisms and hybrid networks (e.g., GT-HAD \cite{lian2024gt}, AdaptHAD \cite{liu2025adaptive}, and KANGT \cite{KANGT}), as well as non-Euclidean modeling like MGAE \cite{MGAE}. More recently, to alleviate computational burdens, the field embraces State Space Models (SSMs) \cite{mamba}, such as CWIMamba \cite{he2025cwimamba} and MMR-HAD \cite{fu2025mmr}, to capture extensive spectral sequences with linear-time complexity.Although these reconstruction-based networks demonstrate robust baseline performance, practical limitations remain in mitigating target memorization and preserving small-scale local details. To address these limitations, our R2VD framework utilizes an advanced full-resolution autoencoder designed specifically for high-fidelity background manifold purification, preventing sub-pixel topological attenuation while suppressing gradient contamination from potential anomalies.

\begin{figure*}[t]
    \centering
    \includegraphics[width=\textwidth]{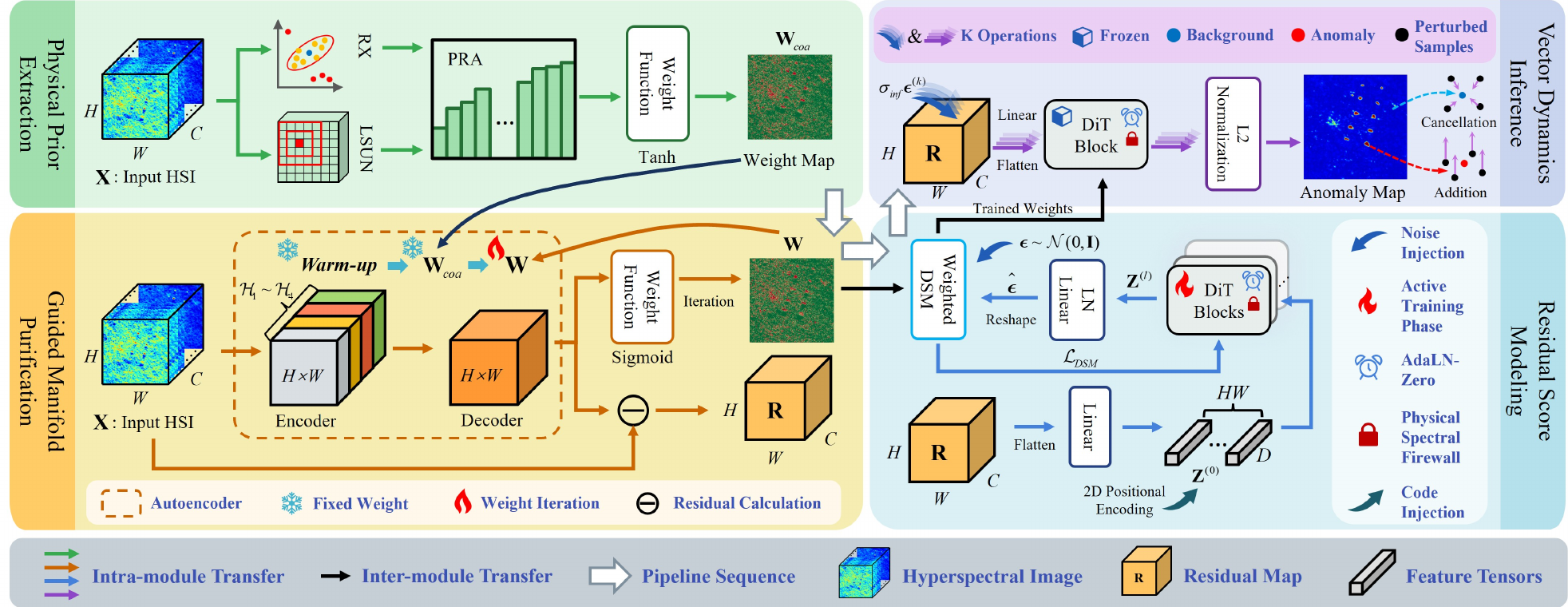}
    \caption{Architecture of the proposed R2VD framework. Physical Prior Extraction: Generates a lenient initial weight map ($\mathbf{W}_{coa}$) from dual-stream detectors (RX and LSUN) via PRA. Guided Manifold Purification: Decouples anomalies using the OCA with dynamic weight iterations, outputting a purified residual map and a weight map ($\mathbf{W}$). Residual Score Modeling: A DiT learns the background score field under the PSF constraint. Vector Dynamics Inference: Injects $K$ perturbations into the frozen DiT field, leveraging the vector cancellation of background gradients and the deterministic vector addition of anomalous gradients to generate the final Anomaly Map.}
    \vspace*{-0.8\baselineskip}
    \label{fig:overview}
\end{figure*}

\subsection{Generative Diffusion and Score Modeling}

Recent HAD research has increasingly explored generative modeling to overcome the limitations of conventional reconstruction based paradigms~\cite{hu2025recent}. Grounded in stochastic differential equations~\cite{song2020score}, Denoising Diffusion Probabilistic Models~\cite{ho2020denoising} and Score Based Generative Models~\cite{song2019generative} learn the score function, namely the gradient of the log density, to characterize the underlying data distribution. The strong representation and generation capabilities of diffusion models have also been demonstrated in image classification~\cite{DiffHC}, hyperspectral fusion~\cite{DiffHSR}, and a broad range of controllable generation tasks. In particular, structured conditions and explicit priors have been exploited for pose guided person generation~\cite{shen2024imagpose}, fine grained garment generation~\cite{shen2025imaggarment}, long term talking face generation~\cite{shenlong}, subject consistent image editing~\cite{shen2025imagedit}, and layout controllable editing~\cite{shen2025imagharmony}. These studies highlight the effectiveness of incorporating task specific priors into diffusion dynamics to improve structural consistency and controllability.
Within HAD, DBD~\cite{DBD} pioneers the integration of diffusion modeling with low rank representation, while BSDM~\cite{ma2025bsdm} models the background as noise to improve anomaly suppression. Subsequent methods introduce spatial and frequency priors, as in FMDiff~\cite{FDiff} and WDHAD~\cite{WDHAD}, or localized dual window modeling, as in DWSDiff~\cite{dwsdiff}, to further enhance anomaly isolation. ScoreAD~\cite{ScoreAD} instead directly employs the learned score function as a detection signal. However, existing methods generally perform generative modeling in entangled spectral spaces and rely heavily on scalar magnitude based anomaly measures, which can become ambiguous under complex illumination and background variations. R2VD addresses these limitations by learning generative score dynamics in a purified residual space with a full resolution DiT. Rather than relying solely on score magnitude, it characterizes the directional consistency of the high dimensional score field, enabling more robust separation between anomalous targets and background responses.

\section{Proposed Method}\label{sec:method} 

\subsection{Overview of the Proposed Framework}

To address the limitations of conventional scalar-based reconstruction and the confirmation bias in deep hyperspectral anomaly detection, we propose the R2VD framework. As illustrated in Fig. \ref{fig:overview}, R2VD conceptualizes anomaly detection as an interactive vector dynamics process, with its overall architecture systematically decoupled into four consecutive stages: PPE, GMP, RSM, and VDI. Initially, the PPE stage processes the input HSI $\mathbf{X} \in \mathbb{R}^{H \times W \times C}$ through dual-stream coarse detectors to provide physical situational awareness, generating a lenient initial physical weight map $\mathbf{W}_{coa} \in \mathbb{R}^{H \times W \times 1}$. Subsequently, the original HSI $\mathbf{X}$ is fed into the GMP stage, which utilizes the full-resolution OCA. Guided initially by $\mathbf{W}_{coa}$ to compute the weighted reconstruction loss, the autoencoder autonomously refines its weights during training, ultimately outputting a converged confidence weight map $\mathbf{W} \in \mathbb{R}^{H \times W \times 1}$ and a purified residual map $\mathbf{R} \in \mathbb{R}^{H \times W \times C}$. In the RSM stage, the residual $\mathbf{R}$ is used to train a full-resolution DiT to estimate the high-dimensional background score field while blocking spectral leakage via the PSF. Finally, the VDI stage injects $K$ random perturbations into $\mathbf{R}$ within the frozen DiT field. By evaluating vector topological consistency, wherein background noise undergoes destructive interference and anomalies trigger constructive alignment, the framework generates the final anomaly map $\mathcal{A} \in \mathbb{R}^{H \times W \times 1}$. Throughout the framework, $E_{\max}^{\mathrm{OCA}}$ and $E_{\max}^{\mathrm{DiT}}$ denote the total training epochs of the OCA and DiT, respectively. The GMP parameters $E_{warm}$ and $E_{update}$ specify the unit-weight warm-up duration and the subsequent weight-map update interval. For residual score modeling, $T$ denotes the number of discrete diffusion noise levels, $\sigma_t$ denotes the noise scale at time-step $t$, and $\sigma_{\min}$ and $\sigma_{\max}$ specify the endpoints of the noise schedule. During VDI, $t_{\mathrm{inf}}$ denotes the fixed inference time step, and $K$ denotes the number of independent perturbations.

\subsection{Physical Prior Extraction}\label{III-B}

The efficacy of HAD is deeply rooted in the robust characterization of background manifolds, where statistical modeling and geometric subspace analysis provide complementary perspectives. Statistical methods capture global distribution outliers, whereas geometric approaches identify local manifold deviations. Given an input HSI $\mathbf{X}$, where $\mathbf{x}_i \in \mathbb{R}^C$ represents the spectral vector at spatial location $i$, the PPE stage establishes a dual-stream coarse detection mechanism to provide the subsequent generative dynamics with physical situational awareness.

Specifically, the first stream implements the RX algorithm \cite{RX} to evaluate the Mahalanobis distance $D_{RX}(\mathbf{x}_i)$ under a multivariate Gaussian assumption:
\vspace*{-0.2\baselineskip}
\begin{equation}
    D_{RX}(\mathbf{x}_i) = (\mathbf{x}_i - \boldsymbol{\mu})^T \boldsymbol{\Sigma}^{-1} (\mathbf{x}_i - \boldsymbol{\mu}).
    \label{RX}
\end{equation}
Simultaneously, the second stream utilizes LSUN \cite{LSUN} to identify the projection residual $D_{LSUN}(\mathbf{x}_i)$ relative to a low-rank background subspace basis $\mathbf{P}^{\perp}$:
\vspace*{-0.2\baselineskip}
\begin{equation}
    D_{LSUN}(\mathbf{x}_i) = ||\mathbf{P}^{\perp} \mathbf{x}_i||_2^2.
    \label{LSUN}
\end{equation}
Since $D_{RX}$ and $D_{LSUN}$ operate in distinct physical dimensions and scales, direct fusion is inherently biased. To achieve scale-invariant integration, we propose a PRA mechanism that transforms raw scores into a rank-based probability space via an empirical cumulative distribution function. For each detector $m \in \{RX, LSUN\}$, the rank-aligned score $R_m(\mathbf{x}_i)$ is formulated as the ascending order index normalized by the total number of pixels $N$, ensuring $R_m \in [0, 1]$. The average consensus is then defined as $P_{avg}(\mathbf{x}_i) = 0.5 ( R_{RX} + R_{LSUN} )$.

To quantitatively guide the subsequent manifold purification, these physical priors are converted into a lenient initial weight map $\mathbf{W}_{coa}$. Let $\tau_{anom}$ be the threshold determined by a predefined anomaly ratio $\eta$. Crucially, rather than acting as a sensitive decision threshold that risks test-set information leakage, $\eta$ is positioned here as a lenient boundary hyperparameter intended only to establish a soft attention window. We normalize the consensus scores as $t_i = P_{avg}(\mathbf{x}_i) / \tau_{anom}$ and implement the piecewise weight function:
\vspace*{-0.2\baselineskip}
\begin{equation}
    \mathbf{W}_{coa}(\mathbf{x}_i) = 
    \begin{cases} 
    \epsilon_{\mathrm{w}}, & t_i \ge 1.0 \\ 
    1.0, & t_i \le \theta_{gap} \\ 
     \mathcal{C}(t_i, \theta_{gap}, k_{coa}) \, & \theta_{gap} < t_i < 1.0
    \end{cases},
    \label{w_coa}
\end{equation}
where $\epsilon_{\mathrm{w}}$ is a small constant for numerical safety, $\theta_{\mathrm{gap}}$ represents the tolerance offset, and $k_{\mathrm{coa}}$ controls the transition slope of the lenient initial weighting function. The function $\mathcal C(\cdot)$ is a shifted hyperbolic tangent function:
$\mathcal{C}(t, \theta, k) = \frac{1}{2} [ 1 - \tanh(k (t - \theta)) ]$. 

\begin{figure}[t]
    \centering
    \includegraphics[width=0.44\textwidth]{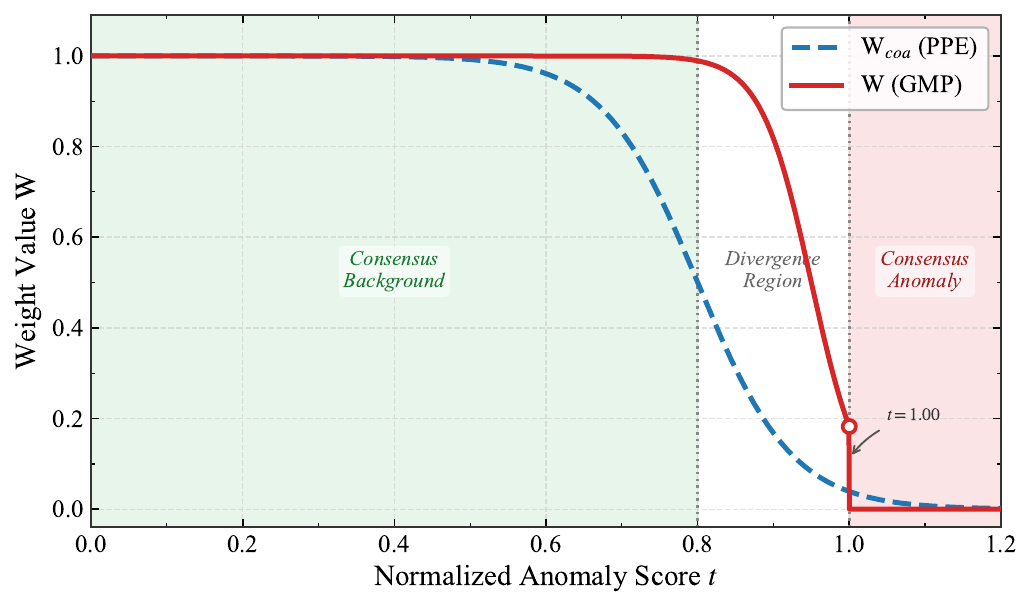}
    \vspace*{-0.5\baselineskip}
    \caption{Illustration of weight function dynamics. The blue dashed curve represents the lenient weight $\mathbf{W}_{coa}$ in the PPE stage to mitigate early confirmation bias. The red solid curve denotes the strict truncation weight $\mathbf{W}$ in the GMP stage, featuring an absolute gradient isolation at $t=1.00$. Background colors partition the score space into consensus background (green), divergence region (white), and consensus anomaly (red).}
    \vspace*{-0.8\baselineskip}
    \label{fig:weight_dynamics}
\end{figure}

It is worth noting that the final detection performance of R2VD is highly robust to the empirical specification of $\eta$. As illustrated in Fig. \ref{fig:weight_dynamics}, the soft-attention boundary outlined by the blue dashed curve deliberately preserves manifold diversity to mitigate early-stage confirmation bias. Any initial alignment errors or blind spots introduced by a misspecified $\eta$ are autonomously rectified during the subsequent GMP stage through the autoencoder's iterative weight self-correction, ensuring entirely unsupervised and stable convergence without requiring dataset-specific tuning.

\subsection{Guided Manifold Purification}\label{III-C}
The GMP stage decouples anomalies from the background manifold while preserving sub-pixel spatial fidelity. Conventional reconstruction methods rely on spatial downsampling, leading to the vanishing of fragile sub-pixel targets. Furthermore, unsupervised optimization without explicit gradient regulation frequently triggers confirmation bias, causing the network to memorize anomalous signatures. To mitigate these spatial and structural limitations, we propose a full-resolution OmniContext Autoencoder (OCA) regulated by a dynamic weight-update strategy.

\begin{figure}[t]
    \centering
    \includegraphics[width=0.44\textwidth]{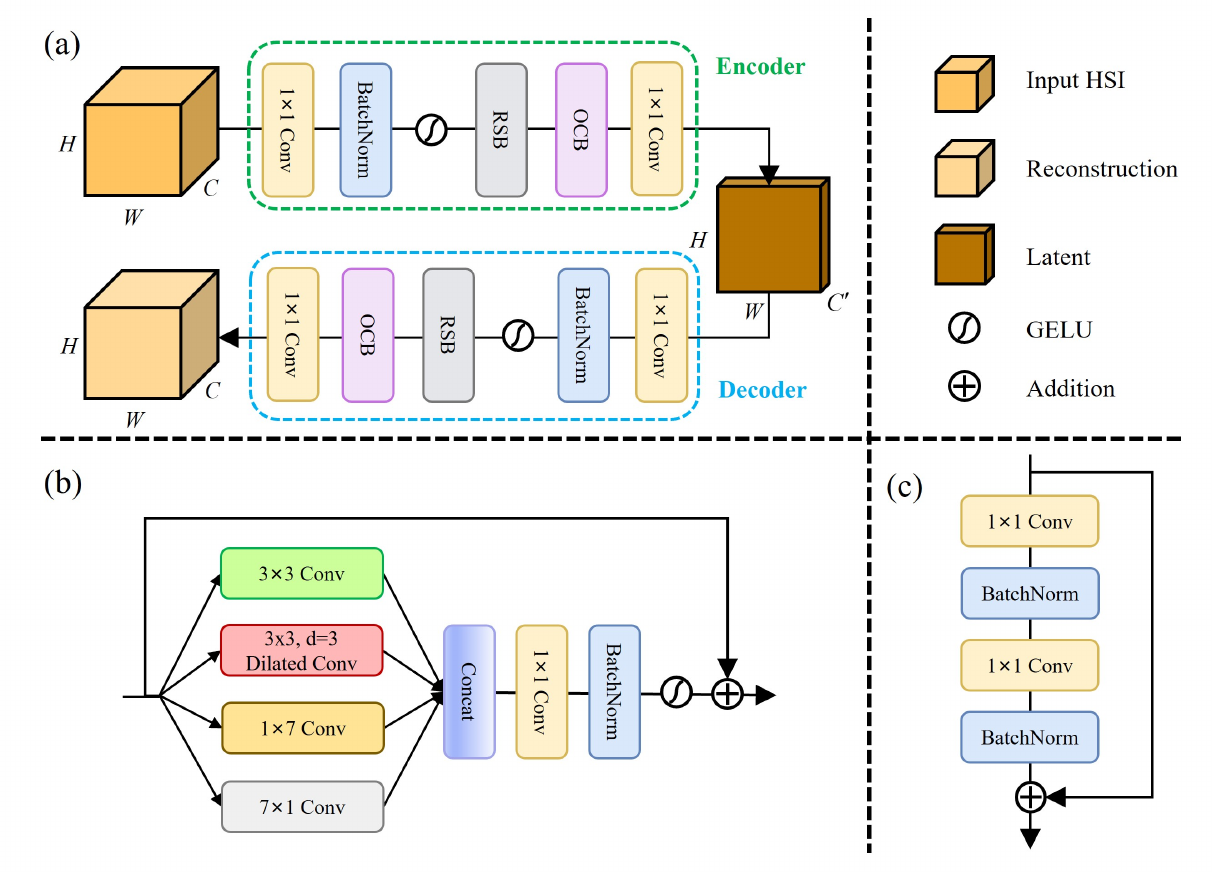} 
    \caption{Architecture of the OCA. (a) The macro-architecture preserves the $H \times W$ spatial dimensions to prevent sub-pixel anomaly vanishing. (b) The OCB extracts multi-scale and directional context features via four parallel branches. (c) The RSB deepens spectral non-linearity using $1 \times 1$ convolutions without spatial blurring.}
    \label{fig:autoencoder}
    \vspace*{-0.8\baselineskip}
\end{figure}

As illustrated in Fig. \ref{fig:autoencoder}(a), the OCA prohibits spatial downsampling to maintain the original $H \times W$ resolution. To expand the receptive field without loss of details, the OmniContext Block (OCB) in Fig. \ref{fig:autoencoder}(b) computes four parallel branches for an intermediate feature map $\mathbf{F}$:
\vspace*{-0.2\baselineskip}
\begin{equation}
    \mathbf{F}_{multi} = \text{Concat}\left( \mathcal{H}_1(\mathbf{F}), \mathcal{H}_2(\mathbf{F}), \mathcal{H}_3(\mathbf{F}), \mathcal{H}_4(\mathbf{F}) \right),
    \label{OCB}
\end{equation}
where $\mathcal{H}_1$ is a standard $3 \times 3$ convolution, $\mathcal{H}_2$ is a $3 \times 3$ dilated convolution, and $\mathcal{H}_3, \mathcal{H}_4$ are $1 \times 7$ and $7 \times 1$ asymmetric convolutions. The fused features yield $\mathbf{F}_{out} = \mathbf{F} + \sigma(\mathcal{B}(\mathcal{P}_{1\times1}(\mathbf{F}_{multi})))$, supplemented by Spectral Residual Blocks (RSBs).

The OCA executes a fully unsupervised, single-scene self-optimization process optimized via a weighted reconstruction loss:
\vspace*{-0.2\baselineskip}
\begin{equation}
    \mathcal{L}^{(e)} = \frac{1}{N} \sum_{i=1}^N \mathbf{W}^{(e)}_i
    \cdot \left\|\mathbf{x}_i-
    [\mathrm{OCA}_{\boldsymbol{\phi}}(\mathbf{X})]_i\right\|_2^2,
    \label{loss}
\end{equation}
where $\mathbf{W}^{(e)} \in \mathbb{R}^{H \times W \times 1}$ is the epoch-driven dynamic weight map. During the initial warm-up phase ($e \le E_{warm}$), we set $\mathbf{W}^{(e)}_i = 1$ to prioritize global background manifold learning.

Upon concluding the warm-up ($e = E_{warm} + 1$), we initialize the physical guidance by setting $\mathbf{W}^{(e)} = \mathbf{W}_{coa}$ from the PPE stage. For the remaining training duration ($e > E_{warm} + 1$), $\mathbf{W}^{(e)}$ is adaptively updated every $E_{update}$ epochs based on the normalized reconstruction error $t_i^{ae} =
\left\|\mathbf{x}_i-
[\mathrm{OCA}_{\boldsymbol{\phi}}(\mathbf{X})]_i\right\|_2^2
/ \tau_{ae}^{(e)}$, where $\tau_{ae}^{(e)}$ denotes the threshold corresponding to the prior anomaly percentile $\eta$. This internal update implements a strict truncation via a steep Sigmoid curve $\mathcal{C}_{ae}(\cdot)$:
\vspace*{-0.2\baselineskip}
\begin{equation}
    \mathbf{W}^{(e)}_i = 
    \begin{cases} 
    0, & t_i^{ae} \ge 1.0 \\ 
    1.0, & t_i^{ae} \le \theta_{gap}^{ae} \\ 
     \mathcal{C}_{ae}(t_i^{ae}, \theta_{gap}^{ae}, k_{ae}), & \theta_{gap}^{ae} < t_i^{ae} < 1.0
    \end{cases},
    \label{w_ae}
\end{equation}
where $\mathcal{C}_{ae}(t, \theta, k) = [1 + \exp(k (t - \theta))]^{-1}$, $\theta_{gap}^{ae}$ represents the strict background tolerance threshold, and $k_{ae}$ controls the scaling slope. Setting $\mathbf{W}^{(e)}_i = 0$ for $t_i^{ae} \ge 1.0$ suppresses the direct reconstruction-loss contribution of high-error pixels, thereby reducing the tendency of the network to memorize anomalous signatures.

This dynamic boundary tightening loop enables autonomous error correction and robust manifold purification. The entire optimization automatically terminates at a fixed maximum epoch $E_{\max}^{\mathrm{OCA}}$ without heuristic post-selection. Ultimately, the GMP stage outputs a converged weight map $\mathbf{W}$ and a purified residual map
$\mathbf{R}=\mathbf{X}-\mathrm{OCA}_{\boldsymbol{\phi}}(\mathbf{X})$.
Before residual score modeling, the residual is standardized as
$\bar{\mathbf{R}}=\mathbf{R}/(\operatorname{std}_{h,w,c}(\mathbf{R})+\xi)$,
where $\operatorname{std}_{h,w,c}(\cdot)$ denotes the standard deviation over all spatial and spectral entries and $\xi>0$ is a numerical stabilization constant.

\subsection{Residual Score Modeling}\label{III-D}
Following the GMP stage, the normalized purified residual space
$\bar{\mathbf{R}}$ exhibits substantially reduced non-Gaussian background interference, providing a refined manifold for high-dimensional probability modeling. To effectively estimate the score field of this data distribution, we introduce a full-resolution DiT. To maintain high spatial fidelity, the architecture refrains from spatial downsampling; each normalized residual vector $\bar{\mathbf r}_i\in\mathbb{R}^{C}$ is directly treated as an individual token. The normalized residual map $\bar{\mathbf{R}}$ is linearly projected into a $D$-dimensional embedding space, yielding a token sequence $\mathbf{Z}^{(0)}\in\mathbb{R}^{(HW)\times D}$ supplemented by 2D sinusoidal positional encodings.

\begin{figure}[t]
    \centering
    \includegraphics[width=0.44\textwidth]{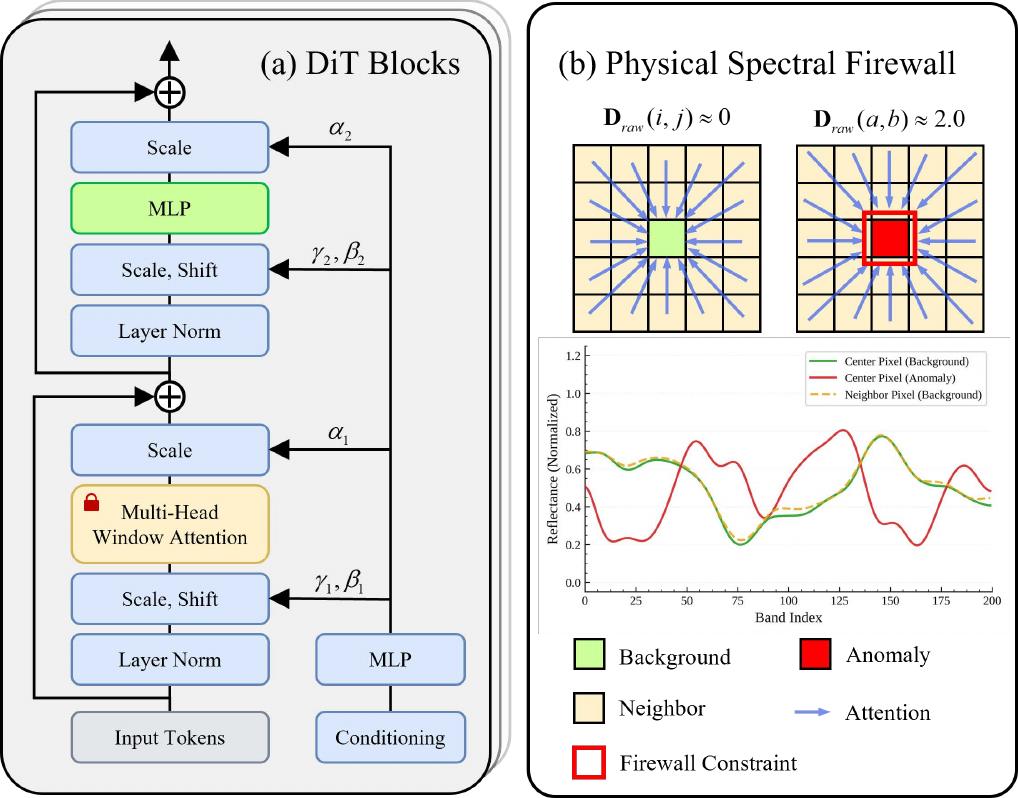}
    \caption{Architecture of the RSM module. (a) The DiT block incorporating AdaLN-Zero modulation for timestep conditioning. (b) The mechanism of the PSF.}
    \label{fig:dit_structure}
    \vspace*{-1.2\baselineskip}
\end{figure}

To condition the network on the discrete diffusion time-step index $t\in\{0,\ldots,T-1\}$, we employ the AdaLN-Zero mechanism \cite{peebles2023scalable}, as illustrated in Fig. \ref{fig:dit_structure}(a). For the $l$-th DiT block, the time step $t$ is mapped via a Multilayer Perceptron to generate modulation parameters $\{\gamma, \beta, \alpha\}$ for latent scaling and shifting. To manage the $O((HW)^2)$ computational complexity, a Window Attention mechanism partitions the latent space into non-overlapping windows of size $win \times win$.

The core innovation is the integration of a PSF within the windowed attention (see Fig. \ref{fig:dit_structure}(b)). Utilizing the original HSI $\mathbf{X}$ as a rigid physical constraint, we compute a spectral distance matrix $\mathbf{D}_{raw} \in \mathbb{R}^{L_{win} \times L_{win}}$ in the $L_2$-normalized spectral space:
\vspace*{-0.2\baselineskip}
\begin{equation}
\mathbf{D}_{raw}(i, j) = \left\| \frac{\mathbf{x}_i}{\|\mathbf{x}_i\|_2} - \frac{\mathbf{x}_j}{\|\mathbf{x}_j\|_2} \right\|_2.
\label{psf}
\end{equation}
Applying $L_2$-normalization establishes a strict geometric equivalence between the Euclidean and Cosine distances, formulated as $\| \frac{\mathbf{x}_i}{\|\mathbf{x}_i\|_2} - \frac{\mathbf{x}_j}{\|\mathbf{x}_j\|_2} \|_2^2 = 2 - 2\cos(\theta_{i,j})$. This bypasses magnitude ambiguity by leveraging robust directional consistency. The attention logit is subsequently penalized by a constant coefficient $\lambda$ to establish the firewall:
\vspace*{-0.2\baselineskip}
\begin{equation}
\mathbf{Z}_{attn} = \text{Softmax}\left( \frac{\mathbf{Q}\mathbf{K}^T}{\sqrt{D_{head}}} - \lambda \cdot \mathbf{D}_{raw} \right)\mathbf{V}.
\label{attn}
\end{equation}
Due to the Softmax function's exponential nature, a large physical distance $\mathbf{D}_{raw}$ prompts the penalty $-\lambda \cdot \mathbf{D}_{raw}$ to effectively attenuate attention weights toward zero. Fixing $\lambda$ as a physical constant circumvents the gradient game paradox where the penalty might otherwise be optimized away. This constraint mitigates cross-spectral feature leakage and encourages the model to estimate score fields within localized spectral manifolds.

We use $T$ discrete noise levels and construct a geometric noise schedule as
\vspace*{-0.2\baselineskip}
\begin{equation}
\sigma_t=
\sigma_{\min}
\left(
\frac{\sigma_{\max}}{\sigma_{\min}}
\right)^{\frac{t}{T-1}},
\quad
t\in\{0,\ldots,T-1\}.
\label{noise_schedule}
\end{equation}
At each DiT training iteration, $t$ is sampled uniformly from $\{0,\ldots,T-1\}$ and $\boldsymbol{\epsilon}\sim\mathcal{N}(\mathbf{0},\mathbf{I})$ is sampled to construct $\bar{\mathbf r}_{t,i}=\bar{\mathbf r}_i+\sigma_t\boldsymbol{\epsilon}_i$. The DiT takes the noisy normalized residual, its corresponding time step, and the original HSI as inputs, and outputs the predicted noise $\hat{\boldsymbol{\epsilon}}=\boldsymbol{\epsilon}_{\boldsymbol{\theta}}(\bar{\mathbf r}_t,t,\mathbf{X})$. Here, $\mathbf{X}$ is used to impose the PSF within window attention, whereas $\mathbf{W}$ is used only to weight the training objective. To ensure the score field characterizes the purified background manifold, we integrate the converged weight map $\mathbf{W}$ from the GMP stage into the training objective. The Weighted Denoising Score Matching (DSM) loss $\mathcal{L}_{DSM}$ is formulated as:
\vspace*{-0.2\baselineskip}
\begin{equation}
    \mathcal{L}_{DSM} =
    \mathbb{E}_{t,\boldsymbol{\epsilon}}
    \left[
    \frac{1}{HW}\sum_{i=1}^{HW}\mathbf{W}_i\cdot
    \left\|
    \boldsymbol{\epsilon}_{\boldsymbol{\theta}}
    (\bar{\mathbf r}_{t,i},t,\mathbf{X})
    -\boldsymbol{\epsilon}_i
    \right\|_2^2
    \right].
    \label{DSM}
\end{equation}
This objective biases score-field learning toward high-confidence background positions by suppressing the direct loss contribution of low-confidence positions. These positions remain in the self-attention computation and may still serve as keys and values for other queries. The PSF therefore provides a complementary constraint: for spectrally dissimilar pixel pairs, the penalty term $-\lambda\mathbf{D}_{raw}$ attenuates their attention connections and reduces the indirect influence of low-confidence anomalous features on background score learning. Thus, $\mathbf{W}$ controls direct DSM supervision, whereas the PSF regulates cross-token information transport. After convergence, RSM outputs the trained noise-prediction model $\boldsymbol{\epsilon}_{\boldsymbol{\theta}}$ rather than an anomaly map; this model is subsequently frozen and used by VDI to obtain score vectors.

\subsection{Vector Dynamics Inference}\label{III-E}
Conventional reconstruction paradigms depend on the scalar $L_2$ norm, which lacks directional awareness and exhibits ambiguity under complex illumination variations or high-frequency textures. To resolve this, the proposed VDI stage shifts the evaluation from magnitude thresholding to probability gradient stability analysis. Here, the full-resolution DiT parameterizes the learned background score field
$\mathbf{S}(\bar{\mathbf r})\equiv
\nabla_{\bar{\mathbf r}}\log p(\bar{\mathbf r})$.

VDI is an inference-only stage that converts the score field learned by RSM into the final anomaly map. It operates on the purified residual using the frozen DiT, while retaining the original HSI $\mathbf{X}$ to preserve the PSF-constrained attention mechanism established during RSM training. No parameter update or additional reconstruction is performed in this stage. For each of the $K$ independent perturbations, VDI fixes the inference time step $t_{\mathrm{inf}}$ and injects Gaussian noise with the corresponding schedule-controlled scale $\sigma_{t_{\mathrm{inf}}}$. The $k$-th perturbed normalized residual vector at pixel $i$ is
$\bar{\mathbf r}_{\mathrm{inf},i}^{(k)}
=\bar{\mathbf r}_i+
\sigma_{t_{\mathrm{inf}}}\boldsymbol{\epsilon}_i^{(k)}$,
where $\boldsymbol{\epsilon}_i^{(k)}
\sim\mathcal{N}(\mathbf{0},\mathbf{I})$. The resulting DiT prediction is converted into a score vector through Tweedie's formula:
\vspace*{-0.2\baselineskip}
\begin{equation}
\mathbf{S}_i^{(k)}
\equiv
\nabla_{\bar{\mathbf r}_{\mathrm{inf},i}^{(k)}}
\log p(\bar{\mathbf r}_{\mathrm{inf},i}^{(k)})
=
-\frac{
\boldsymbol{\epsilon}_{\boldsymbol{\theta}}
(\bar{\mathbf r}_{\mathrm{inf},i}^{(k)},
t_{\mathrm{inf}},\mathbf{X})
}{
\sigma_{t_{\mathrm{inf}}}
}.
\label{score}
\end{equation}
To eliminate magnitude scale bias, strict $L_2$-normalization is applied to extract the pure directional score unit vector:
\vspace*{-0.2\baselineskip}
\begin{equation}
    \mathbf{u}_i^{(k)} = \frac{\mathbf{S}_i^{(k)}}{\|\mathbf{S}_i^{(k)}\|_2 + \xi},
    \label{u}
\end{equation}
where $\xi$ is a small stabilization constant. The final anomaly metric is derived from the cumulative trajectory vector:
\vspace*{-0.2\baselineskip}
\begin{equation}
    \mathbf{v}_{cum, i} = \sum_{k=1}^K \mathbf{u}_i^{(k)}.
    \label{vector}
\end{equation}

We provide a rigorous probabilistic justification for this formulation based on score-field convergence boundaries. For an in-distribution background pixel $\bar{\mathbf r}_i$, the purified sample already resides within the high-probability density basin of the background manifold $\mathcal{M}$. When perturbed by isotropic micro-noise $\sigma_{t_{\mathrm{inf}}} \boldsymbol{\epsilon}_i^{(k)}$, the sample disperses omnidirectionally around the local mode. Because the score field $\mathbf{S}(\bar{\mathbf r})$ represents the steepest ascent vector pointing toward the local manifold ridge, the generated unit vectors $\mathbf{u}_i^{(k)}$ track these symmetrical local recovery paths. Under $K$ independent stochastic samplings, these vectors behave as omnidirectional uniform variables over the unit hypersphere, yielding an expected value of $\mathbb{E}[\mathbf{u}_i^{(k)}] = \mathbf{0}$. Consequently, the random dispersion triggers a vector cancellation effect, driving the expected cumulative norm toward zero:
\vspace*{-0.2\baselineskip}
\begin{equation}
    \lim_{K \to \infty} \frac{1}{K}\left\| \mathbf{v}_{cum, i} \right\|_2 = 0, \quad \forall \bar{\mathbf r}_i \in \mathcal{M}.
    \label{bg_proof}
\end{equation}

Conversely, for an out-of-distribution anomaly pixel $\bar{\mathbf r}_a$, its coordinates reside in a low probability density region far from $\mathcal{M}$, explicitly isolated by the weight map $\mathbf{W}$ and the PSF. The macroscopic distance from $\bar{\mathbf r}_a$ to the nearest background manifold mode is much larger than the injected microscopic perturbation scale ($dist(\bar{\mathbf r}_a, \mathcal{M}) \gg \sigma_{t_{\mathrm{inf}}}$). Therefore, the micro-noise $\sigma_{t_{\mathrm{inf}}}\boldsymbol{\epsilon}_a^{(k)}$ shifts the sample coordinates negligibly relative to the global manifold landscape. Tweedie's projection vector under SDE constraints persistently points along a uniform global gradient direction toward the distant background cluster across $K$ independent stochastic samplings. This strong directional alignment forces the unit vectors to converge toward a uniform deterministic direction $\mathbf{u}_a^*$, executing a vector addition effect that drives the cumulative norm to its theoretical maximum $K$:
\vspace*{-0.2\baselineskip}
\begin{equation}
    \left\| \mathbf{v}_{cum, a} \right\|_2 \approx \left\| \sum_{k=1}^K \mathbf{u}_a^* \right\|_2 = K, \quad \forall \bar{\mathbf r}_a \notin \mathcal{M}.
    \label{anomaly_proof}
\end{equation}

The final Anomaly Map $\mathcal{A}$ is obtained via spatial min-max normalization:
\vspace*{-0.2\baselineskip}
\begin{equation}
    \mathcal{A}_i = \frac{\|\mathbf{v}_{cum, i}\|_2 - \min_j (\|\mathbf{v}_{cum, j}\|_2)}{\max_j (\|\mathbf{v}_{cum, j}\|_2) - \min_j (\|\mathbf{v}_{cum, j}\|_2) + \xi}.
    \label{anomaly}
\end{equation}
Thus, VDI uses the cumulative norm of normalized score vectors, rather than a scalar reconstruction error, as its pre-normalization anomaly response. Background perturbations tend to yield mutually cancelling directions, whereas anomalous perturbations yield consistently aligned directions. The normalized map $\mathcal{A}$ is the sole output of VDI and the final detection result of R2VD. The data flow and interactions among PPE, GMP, RSM, and VDI are summarized in Algorithm \ref{algo:r2vd}.

For clarity, R2VD shares the score-vector foundation with ScoreAD \cite{ScoreAD}, but differs in its modeling space and inference objective. ScoreAD directly converts a score estimate in the original HSI distribution into a single scalar anomaly response, where background and anomalous signatures may remain entangled. In contrast, R2VD establishes a PSF-constrained score field in the GMP-purified residual space and uses the score vectors as directional evidence. VDI aggregates normalized directions obtained from $K$ independent perturbations, so that incoherent background directions tend to cancel whereas consistently oriented anomalous directions are accumulated. The final decision is therefore based on directional consistency across perturbations rather than on a single score magnitude.

\begin{algorithm}[t]
\caption{The Proposed R2VD Framework}
\label{algo:r2vd}
\begin{algorithmic}[1]
\Statex \textbf{Input:} HSI $\mathbf{X}$; hyperparameters
$K$, $\lambda$, $\eta$, $T$, $\sigma_{\min}$, $\sigma_{\max}$,
$t_{\mathrm{inf}}$, $E_{\max}^{\mathrm{OCA}}$,
$E_{\max}^{\mathrm{DiT}}$, $E_{warm}$,
$E_{update}$, and $\xi$
\Statex \textbf{Output:} anomaly map $\mathcal{A}$
\Statex {\color{gray}\vspace{-0.2cm}\rule{\linewidth}{0.4pt}}
\State \textbf{PPE:} Compute RX and LSUN scores from $\mathbf{X}$
\State $\mathbf{W}_{\mathrm{coa}}\leftarrow
\textsc{PRA-Weighting}(\mathrm{RX},\mathrm{LSUN},\eta)$
\Statex {\color{gray}\vspace{-0.2cm}\rule{\linewidth}{0.4pt}}
\State \textbf{GMP:} Initialize OCA parameters $\boldsymbol{\phi}$ and
$\mathbf{W}\leftarrow\mathbf{W}_{\mathrm{coa}}$
\For{$e=1$ \textbf{to} $E_{\max}^{\mathrm{OCA}}$}
    \State $\mathbf{W}_{\mathrm{train}}\leftarrow\mathbf{1}$ if
    $e\leq E_{warm}$; otherwise
    $\mathbf{W}_{\mathrm{train}}\leftarrow\mathbf{W}$
    \State $\widehat{\mathbf{X}}\leftarrow
    \mathrm{OCA}_{\boldsymbol{\phi}}(\mathbf{X})$
    \State Update $\boldsymbol{\phi}$ using Eq.~\eqref{loss}
    with $\mathbf{W}_{\mathrm{train}}$
    \If{$e>E_{warm}$ and $e\bmod E_{update}=0$}
        \State Update $\mathbf{W}$ from the current residual using
        Eq.~\eqref{w_ae}
    \EndIf
\EndFor
\State $\mathbf{R}\leftarrow\mathbf{X}-\widehat{\mathbf{X}}$,
$\bar{\mathbf{R}}\leftarrow
\mathbf{R}/(\operatorname{std}_{h,w,c}(\mathbf{R})+\xi)$
\Statex {\color{gray}\vspace{-0.2cm}\rule{\linewidth}{0.4pt}}
\State \textbf{RSM:} Construct the geometric noise schedule
$\{\sigma_t\}_{t=0}^{T-1}$ using Eq.~\eqref{noise_schedule}
\State Compute $\mathbf{D}_{raw}$ from $\mathbf{X}$
for PSF-constrained attention
\For{$e=1$ \textbf{to} $E_{\max}^{\mathrm{DiT}}$}
    \State Sample $t\sim\mathcal{U}\{0,\ldots,T-1\}$ and
    $\boldsymbol{\epsilon}\sim\mathcal{N}(\mathbf{0},\mathbf{I})$
    \State $\bar{\mathbf{R}}_t\leftarrow
    \bar{\mathbf{R}}+\sigma_t\boldsymbol{\epsilon}$
    \State $\hat{\boldsymbol{\epsilon}}\leftarrow
    \mathrm{DiT}_{\boldsymbol{\theta}}(\bar{\mathbf{R}}_t,t,\mathbf{X})$
    \State Update $\boldsymbol{\theta}$ using Eq.~\eqref{DSM}
\EndFor
\Statex {\color{gray}\vspace{-0.2cm}\rule{\linewidth}{0.4pt}}
\State \textbf{VDI:} Freeze $\boldsymbol{\phi}$ and $\boldsymbol{\theta}$;
set $\mathbf{v}_{\mathrm{cum}}\leftarrow\mathbf{0}$
\For{$k=1$ \textbf{to} $K$}
    \State Sample $\boldsymbol{\epsilon}^{(k)}
    \sim\mathcal{N}(\mathbf{0},\mathbf{I})$
    \State $\bar{\mathbf{R}}_{\mathrm{inf}}^{(k)}
    \leftarrow\bar{\mathbf{R}}+
    \sigma_{t_{\mathrm{inf}}}\boldsymbol{\epsilon}^{(k)}$
    \State Obtain $\mathbf{u}^{(k)}$ from
    $\mathrm{DiT}_{\boldsymbol{\theta}}
    (\bar{\mathbf{R}}_{\mathrm{inf}}^{(k)},
    t_{\mathrm{inf}},\mathbf{X})$
    using Eqs.~\eqref{score} and \eqref{u}
    \State $\mathbf{v}_{\mathrm{cum}}\leftarrow
    \mathbf{v}_{\mathrm{cum}}+\mathbf{u}^{(k)}$
\EndFor
\State $\mathcal{A}\leftarrow
\textsc{MinMax}(\|\mathbf{v}_{\mathrm{cum}}\|_2)$
using Eq.~\eqref{anomaly}
\State \Return $\mathcal{A}$
\end{algorithmic}
\end{algorithm}

\section{Experiment and Analysis}\label{sec:exp}  

\subsection{Experimental Settings}
\subsubsection{Datasets}

We evaluate the proposed R2VD on eight datasets that represent various environmental scenarios. The detailed quantitative characteristics of these datasets, including spatial dimensions, spectral bands, and anomaly ratios, are summarized in Table \ref{tab:dataset_info}. The experimental testbed is categorized into small-scale and large-scale scenarios to assess performance across different data volumes. The small-scale group consists of five datasets. Specifically, Urban-2 \cite{abu} captures urban building clusters, and Aviris-2 \cite{Aviris} features an airport environment. HAD100-95 \cite{li2023you} represents vegetation-dominated mixed areas, while HAD100-40 \cite{li2023you} focuses on the separation of shadows and anomalies within relatively pure backgrounds. Additionally, Hyperion \cite{Hyperion} covers agricultural land surfaces. In addition to the small-scale cases, three large-scale datasets are included to provide a more rigorous evaluation. Cri \cite{Cri} features small rocks scattered across dense vegetation, Segundo \cite{Segundo} portrays large-scale industrial facilities, and UHAD-I \cite{liu2025adaptive} involves the detection of vehicles on road networks.

\begin{table}[t]
\caption{Detailed Information of the Eight Hyperspectral Datasets.}
\vspace*{-0.2\baselineskip}
\centering
\label{tab:dataset_info}
\resizebox{\columnwidth}{!}{
\begin{tabular}{lcccc}
\toprule
Dataset Name & Spatial Size & Bands & Anomaly Pixels & Ratio (\%) \\
\midrule
Urban-2 \cite{abu} & 100 $\times$ 100 & 207 & 155  & 1.55 \\
Aviris-2 \cite{Aviris} & 128 $\times$ 128 & 189 & 120  & 0.73 \\
Cri \cite{Cri} & 400 $\times$ 400 & 46  & 860  & 0.54 \\
HAD100-40 \cite{li2023you} & 64 $\times$ 64   & 276 & 67   & 1.64 \\
HAD100-95 \cite{li2023you} & 64 $\times$ 64   & 276 & 85   & 2.08 \\
Hyperion \cite{Hyperion} & 100 $\times$ 100 & 145 & 32   & 0.32 \\
Segundo \cite{Segundo}     & 250 $\times$ 300 & 224 & 2048 & 2.73 \\
UHAD-I \cite{liu2025adaptive}    & 250 $\times$ 250 & 61  & 302  & 0.48 \\
\bottomrule
\end{tabular}
}
\vspace*{-0.8\baselineskip}
\end{table}

\subsubsection{Comparison Methods}

To comprehensively evaluate R2VD, we compare it with ten representative anomaly detection algorithms across three mainstream paradigms. The first paradigm comprises traditional mathematical models such as the statistical benchmark RX \cite{RX} and the subspace-based LSUN \cite{LSUN}. The second paradigm involves deep reconstruction networks. In addition to the autoencoder-based Auto-AD \cite{wang2021auto}, GT-HAD \cite{lian2024gt}, and guided autoencoder detector GAED \cite{gaed}, this paradigm includes recent self-supervised blind-spot reconstruction methods, namely PDBSNet \cite{PDBSNet}, PUNNet \cite{PUNNet}, and NL2Net \cite{NL2Net}. Specifically, PDBSNet employs pixel-shuffle downsampling to enhance blind-spot background reconstruction, while PUNNet and NL2Net further improve contextual background modeling through global feature injection and coupled local--non-local feature learning, respectively. The third paradigm focuses on recently emerged diffusion models, specifically BSDM \cite{ma2025bsdm} and ScoreAD \cite{ScoreAD}. Furthermore, to demonstrate the modular compatibility of our framework, we conduct extensive tests by feeding the residual representations generated by Auto-AD and GT-HAD into the proposed RSM and VDI stages. These fusion variants are denoted as (Auto-AD)$_{\text{VD}}$ and (GT-HAD)$_{\text{VD}}$, respectively. This approach allows us to evaluate the efficacy of vector dynamics inference when initialized with different manifold purification priors. Finally, we optimize all baseline hyperparameters following their original publications to ensure a fair comparison.

\subsubsection{Evaluation Metrics}

Let $a_i\in[0,1]$ denote the normalized anomaly score of pixel $i$, and let $y_i\in\{0,1\}$ denote its ground-truth label, where $y_i=1$ represents an anomaly and $y_i=0$ represents background. The numbers of anomaly and background pixels are defined as
\begin{equation}
N_1=\sum_{i=1}^{N}y_i,\qquad
N_0=\sum_{i=1}^{N}(1-y_i).
\label{metric_numbers}
\end{equation}

For a decision threshold $\tau\in[0,1]$, the detection probability and false-alarm probability are defined as
\begin{equation}
P_D(\tau)=\frac{1}{N_1}\sum_{i=1}^{N}y_i\mathbb{I}(a_i\geq\tau),
\label{metric_pd}
\end{equation}
\begin{equation}
P_F(\tau)=\frac{1}{N_0}\sum_{i=1}^{N}(1-y_i)\mathbb{I}(a_i\geq\tau),
\label{metric_pf}
\end{equation}
where $\mathbb{I}(\cdot)$ denotes the indicator function. Based on $P_D$, $P_F$, and $\tau$, the reported metrics are defined as follows.

\begin{enumerate}
\item $AUC_{(D,F)}$ measures the overall anomaly--background separability through the conventional ROC curve:
\begin{equation}
AUC_{(D,F)}=\int_{0}^{1}P_D(P_F)\,\mathrm{d}P_F.
\label{metric_auc_df}
\end{equation}

\item $AUC_{(D,\tau)}$ measures target enhancement over varying decision thresholds:
\begin{equation}
AUC_{(D,\tau)}=\int_{0}^{1}P_D(\tau)\,\mathrm{d}\tau.
\label{metric_auc_dt}
\end{equation}

\item $AUC_{(F,\tau)}$ quantifies the false-alarm response of background pixels over varying thresholds:
\begin{equation}
AUC_{(F,\tau)}=\int_{0}^{1}P_F(\tau)\,\mathrm{d}\tau.
\label{metric_auc_ft}
\end{equation}

\item $AUC_{OD}$ jointly evaluates detection capability, target enhancement, and background suppression:
\begin{equation}
AUC_{OD}=AUC_{(D,F)}+AUC_{(D,\tau)}-AUC_{(F,\tau)}.
\label{metric_auc_od}
\end{equation}

\item $AUC_{SNPR}$ measures the target-response advantage relative to background false alarms:
\begin{equation}
AUC_{SNPR}=
\frac{AUC_{(D,\tau)}}
{AUC_{(F,\tau)}+\epsilon_{\mathrm{num}}},
\label{metric_auc_snpr}
\end{equation}
\end{enumerate}

Here, $\epsilon_{\mathrm{num}}$ is a small positive constant introduced to avoid division by zero. All integrals are numerically calculated from the empirical ROC samples using the trapezoidal rule. Larger values are preferred for $AUC_{(D,F)}$, $AUC_{(D,\tau)}$, $AUC_{OD}$, and $AUC_{SNPR}$, whereas a smaller $AUC_{(F,\tau)}$ indicates stronger background suppression.

\subsubsection{Implementation Details}

We conduct all experiments using PyTorch on a single NVIDIA GeForce RTX 5070 Ti Laptop GPU. The optimization of R2VD proceeds in two stages. First, the OCA is trained for $E_{\max}^{\mathrm{OCA}}=100$ epochs using Adam with a learning rate of $0.001$. The first $E_{warm}=3$ epochs use unit reconstruction weights, after which the GMP weight map is updated every $E_{update}=10$ epochs. Second, the full-resolution DiT is trained for $E_{\max}^{\mathrm{DiT}}=1000$ epochs using AdamW with a learning rate of $0.0002$ and a weight decay of $10^{-5}$. A separate linear learning-rate warm-up is applied during the first $50$ DiT optimization iterations.

For diffusion training, we use $T=1000$ discrete noise levels with $\sigma_{\min}=0.01$ and $\sigma_{\max}=2.0$ under the geometric schedule in Eq.~\eqref{noise_schedule}. At each DiT training iteration, the time-step index is sampled uniformly from $\{0,\ldots,999\}$. For the primary R2VD framework, $t_{\mathrm{inf}}=980$, which corresponds to $\sigma_{t_{\mathrm{inf}}}\approx1.8083$; for (Auto-AD)$_{\mathrm{VD}}$ and (GT-HAD)$_{\mathrm{VD}}$, $t_{\mathrm{inf}}=200$, corresponding to $\sigma_{t_{\mathrm{inf}}}\approx0.0289$.

The DiT uses a local attention window size of $8$, a depth of $4$, and an embedding dimension of $128$. The remaining core hyperparameters are fixed as $K=50$, $\lambda=5.0$, and $\eta=0.02$. The slope parameters $k_{\mathrm{coa}}$ and $k_{\mathrm{ae}}$ are empirically fixed at $10$ and $40$, respectively, for all datasets, without scene-specific tuning.

\subsection{Experiment Result} 
\subsubsection{Comparisons of AUCs}

\begin{figure*}[t]
    \centering
    \includegraphics[width=\textwidth]{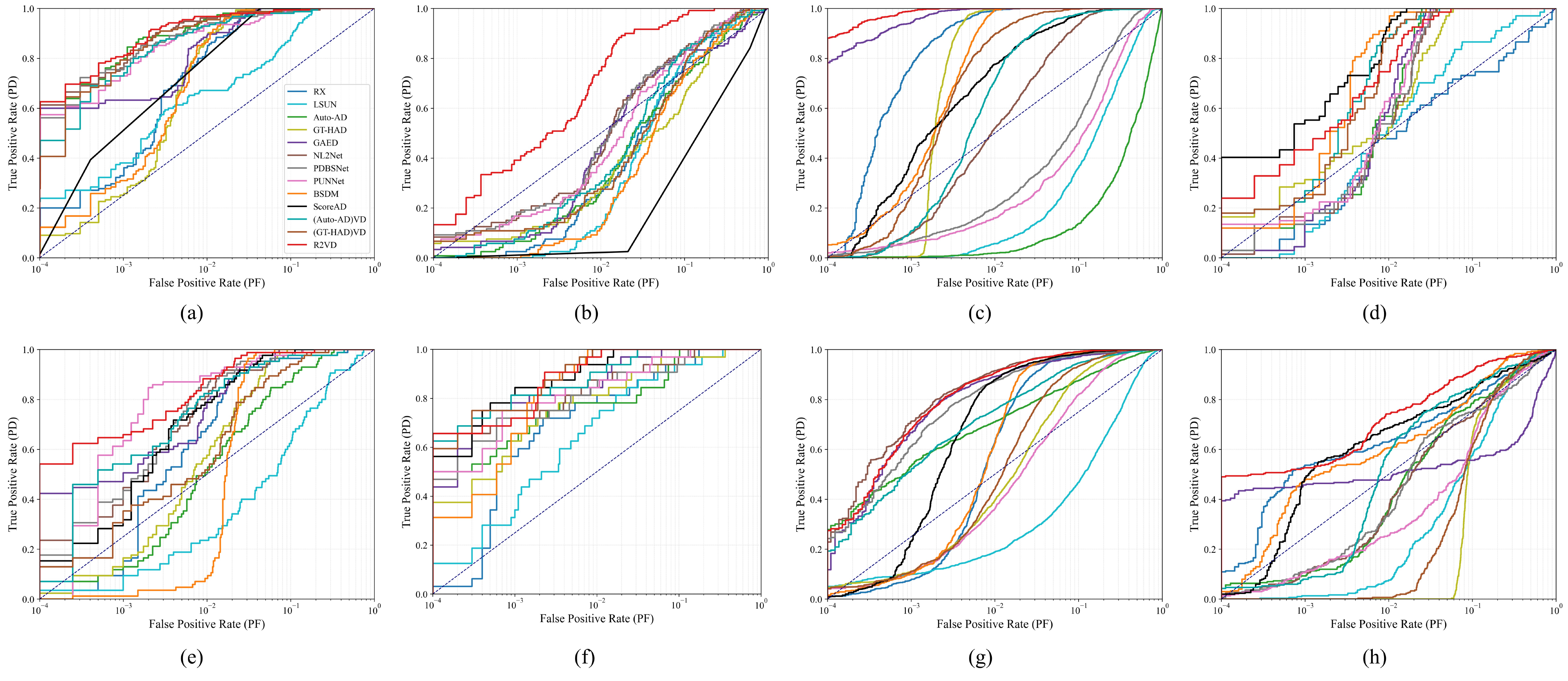}
    \vspace*{-2em}
    \caption{2D ROC curves of the proposed R2VD and other state-of-the-art anomaly detection methods across eight hyperspectral datasets. (a) Urban-2. (b) Aviris-2. (c) Cri. (d) HAD100-40. (e) HAD100-95. (f) Hyperion. (g) Segundo. (h) UHAD-I. In most evaluated datasets, R2VD maintains a stable position in the upper-left region.}
    \vspace*{-0.8\baselineskip}
    \label{fig:visual_roc}
\end{figure*}

The proposed R2VD demonstrates superior background suppression and target detectability across diverse hyperspectral scenes. Operating under unified physical configurations, quantitative evaluations in Table \ref{tab:all_metrics_combined} reveal that R2VD achieves the highest primary $AUC_{(D,F)}$ on six out of the eight datasets. In terms of average performance, our framework secures first place in three critical metrics including $AUC_{(D,F)}$ at $0.9915$, $AUC_{OD}$ at $1.5748$, and $AUC_{SNPR}$ at $127.1535$, alongside a second-place ranking in $AUC_{(D,\tau)}$ at $0.6408$. For complex scenes, R2VD yields a top primary metric of $0.9878$ on Aviris-2 and an exceptional $AUC_{SNPR}$ of $943.8$ on Segundo. This robust detectability is visually corroborated by the two dimensional ROC curves in Fig. \ref{fig:visual_roc}, where R2VD consistently envelopes the upper left corner even under stringent low false alarm conditions.

Furthermore, the integration of RSM and VDI significantly enhances traditional paradigms. By replacing the standard scalar residual evaluation with our RSM and VDI stages, the (Auto-AD)$_{\text{VD}}$ and (GT-HAD)$_{\text{VD}}$ variants exhibit consistent metric improvements over their baseline autoencoder models across most datasets. Specifically, the primary $AUC_{(D,F)}$ of GT-HAD on the Hyperion dataset increases from $0.9823$ to $0.9990$, while the identical metric for Auto-AD on the Cri dataset leaps from $0.5494$ to $0.9850$. Compared with ScoreAD, the complete R2VD framework achieves substantial average improvements across all three metrics, increasing $AUC_{(D,F)}$ from $0.9380$ to $0.9915$, $AUC_{OD}$ from $1.1822$ to $1.5748$, and $AUC_{SNPR}$ from $3.1626$ to $127.1535$. These results demonstrate the overall effectiveness of the proposed R2VD framework relative to the score-based baseline. This consistent performance improvement supports the effectiveness of shifting from scalar reconstruction residuals to the proposed vector inference paradigm.

\begin{table*}[!t]
\centering
\caption{Quantitative results on eight datasets. \textcolor{red}{Red}, \textcolor{darkgreen}{dark green}, and \textcolor{blue}{blue} denote the first, second, and third results, respectively, ranked across all methods.}
\label{tab:all_metrics_combined}
\scriptsize
\setlength{\tabcolsep}{1.2pt}
\renewcommand{\arraystretch}{1.04}
\resizebox{\textwidth}{!}{%
\begin{tabular}{ll*{13}{c}}
\toprule
Dataset & Metric & \shortstack[c]{RX\\\cite{RX}} & \shortstack[c]{LSUN\\\cite{LSUN}} & \shortstack[c]{Auto-AD\\\cite{wang2021auto}} & \shortstack[c]{GT-HAD\\\cite{lian2024gt}} & \shortstack[c]{GAED\\\cite{gaed}} & \shortstack[c]{NL2Net\\\cite{NL2Net}} & \shortstack[c]{PDBSNet\\\cite{PDBSNet}} & \shortstack[c]{PUNNet\\\cite{PUNNet}} & \shortstack[c]{BSDM\\\cite{ma2025bsdm}} & \shortstack[c]{ScoreAD\\\cite{ScoreAD}} & \shortstack[c]{Auto-AD\\+VD} & \shortstack[c]{GT-HAD\\+VD} & \textbf{R2VD} \\
\midrule
\multirow[c]{5}{*}{\shortstack[c]{Urban-2\\\cite{abu}}} & $AUC_{(D,F)}\uparrow$ & 0.9946 & 0.9721 & \textcolor{blue}{\textbf{0.9970}} & 0.9953 & 0.9958 & 0.9969 & 0.9960 & 0.9951 & 0.9964 & 0.9867 & 0.9950 & \textcolor{darkgreen}{\textbf{0.9974}} & \textcolor{red}{\textbf{0.9975}} \\
 & $AUC_{(D,\tau)}\uparrow$ & 0.1178 & 0.0653 & 0.2666 & 0.2183 & 0.1098 & 0.2718 & 0.1842 & 0.2317 & 0.1649 & \textcolor{red}{\textbf{0.9577}} & \textcolor{blue}{\textbf{0.3540}} & 0.2989 & \textcolor{darkgreen}{\textbf{0.7891}} \\
 & $AUC_{(F,\tau)}\downarrow$ & 0.0135 & 0.0020 & 0.0031 & 0.0081 & 0.0045 & \textcolor{darkgreen}{\textbf{0.0007}} & \textcolor{blue}{\textbf{0.0011}} & \textcolor{red}{\textbf{0.0007}} & 0.0133 & 0.7395 & 0.0244 & 0.0182 & 0.0471 \\
 & $AUC_{OD}\uparrow$ & 1.0989 & 1.0354 & 1.2605 & 1.2055 & 1.1011 & 1.2680 & 1.1790 & 1.2262 & 1.1480 & 1.2049 & \textcolor{darkgreen}{\textbf{1.3247}} & \textcolor{blue}{\textbf{1.2782}} & \textcolor{red}{\textbf{1.7395}} \\
 & $AUC_{SNPR}\uparrow$ & 8.7151 & 32.4981 & 86.0995 & 27.0674 & 24.3536 & \textcolor{red}{\textbf{372.9856}} & \textcolor{blue}{\textbf{163.9466}} & \textcolor{darkgreen}{\textbf{352.3328}} & 12.3745 & 1.2951 & 14.5160 & 16.4608 & 16.7709 \\
\midrule
\multirow[c]{5}{*}{\shortstack[c]{Aviris-2\\\cite{Aviris}}} & $AUC_{(D,F)}\uparrow$ & 0.9181 & 0.9150 & 0.8950 & 0.8968 & 0.9034 & \textcolor{blue}{\textbf{0.9389}} & \textcolor{darkgreen}{\textbf{0.9453}} & 0.9365 & 0.8967 & 0.6207 & 0.9241 & 0.9228 & \textcolor{red}{\textbf{0.9878}} \\
 & $AUC_{(D,\tau)}\uparrow$ & 0.1185 & 0.1932 & 0.0850 & 0.2190 & 0.1943 & 0.1733 & 0.1782 & 0.1740 & 0.0483 & \textcolor{red}{\textbf{0.9810}} & 0.1919 & \textcolor{blue}{\textbf{0.3269}} & \textcolor{darkgreen}{\textbf{0.5685}} \\
 & $AUC_{(F,\tau)}\downarrow$ & 0.0386 & 0.0789 & \textcolor{blue}{\textbf{0.0075}} & 0.0301 & 0.0109 & \textcolor{red}{\textbf{0.0059}} & \textcolor{darkgreen}{\textbf{0.0067}} & 0.0079 & 0.0105 & 0.9708 & 0.0450 & 0.0726 & 0.0558 \\
 & $AUC_{OD}\uparrow$ & 0.9980 & 1.0293 & 0.9726 & 1.0857 & 1.0869 & 1.1062 & \textcolor{blue}{\textbf{1.1168}} & 1.1027 & 0.9344 & 0.6309 & 1.0710 & \textcolor{darkgreen}{\textbf{1.1771}} & \textcolor{red}{\textbf{1.5006}} \\
 & $AUC_{SNPR}\uparrow$ & 3.0728 & 2.4478 & 11.4059 & 7.2758 & 17.9101 & \textcolor{red}{\textbf{29.2026}} & \textcolor{darkgreen}{\textbf{26.7171}} & \textcolor{blue}{\textbf{22.0906}} & 4.5831 & 1.0105 & 4.2601 & 4.5010 & 10.1961 \\
\midrule
\multirow[c]{5}{*}{Cri \cite{Cri}} & $AUC_{(D,F)}\uparrow$ & \textcolor{blue}{\textbf{0.9990}} & 0.7782 & 0.5494 & 0.9977 & \textcolor{darkgreen}{\textbf{0.9998}} & 0.9734 & 0.8757 & 0.8285 & 0.9974 & 0.9872 & 0.9850 & 0.9951 & \textcolor{red}{\textbf{0.9999}} \\
 & $AUC_{(D,\tau)}\uparrow$ & 0.1073 & 0.3088 & 0.0409 & 0.0054 & \textcolor{blue}{\textbf{0.4260}} & 0.1486 & 0.1089 & 0.1466 & 0.2813 & \textcolor{darkgreen}{\textbf{0.9384}} & 0.3556 & 0.0842 & \textcolor{red}{\textbf{0.9394}} \\
 & $AUC_{(F,\tau)}\downarrow$ & 0.0237 & 0.2289 & 0.0335 & \textcolor{red}{\textbf{0.0011}} & \textcolor{darkgreen}{\textbf{0.0115}} & 0.0267 & 0.0285 & 0.0482 & 0.0526 & 0.8390 & 0.1033 & \textcolor{blue}{\textbf{0.0169}} & 0.0916 \\
 & $AUC_{OD}\uparrow$ & 1.0826 & 0.8581 & 0.5568 & 1.0020 & \textcolor{darkgreen}{\textbf{1.4142}} & 1.0953 & 0.9561 & 0.9269 & 1.2261 & 1.0866 & \textcolor{blue}{\textbf{1.2373}} & 1.0624 & \textcolor{red}{\textbf{1.8477}} \\
 & $AUC_{SNPR}\uparrow$ & 4.5296 & 1.3488 & 1.2235 & 4.9419 & \textcolor{red}{\textbf{36.9117}} & \textcolor{blue}{\textbf{5.5679}} & 3.8256 & 3.0410 & 5.3436 & 1.1185 & 3.4429 & 4.9904 & \textcolor{darkgreen}{\textbf{10.2516}} \\
\midrule
\multirow[c]{5}{*}{\shortstack[c]{HAD100-40\\\cite{li2023you}}} & $AUC_{(D,F)}\uparrow$ & 0.8651 & 0.9406 & 0.9903 & 0.9860 & 0.9906 & 0.9883 & 0.9888 & 0.9901 & \textcolor{darkgreen}{\textbf{0.9966}} & \textcolor{red}{\textbf{0.9970}} & \textcolor{blue}{\textbf{0.9952}} & 0.9947 & 0.9936 \\
 & $AUC_{(D,\tau)}\uparrow$ & 0.3588 & 0.2023 & 0.2718 & 0.3982 & \textcolor{blue}{\textbf{0.4393}} & 0.3063 & 0.3090 & 0.3204 & 0.2598 & \textcolor{red}{\textbf{0.5297}} & 0.3998 & \textcolor{darkgreen}{\textbf{0.4713}} & 0.3946 \\
 & $AUC_{(F,\tau)}\downarrow$ & 0.1394 & 0.0864 & 0.0244 & 0.0280 & \textcolor{darkgreen}{\textbf{0.0151}} & \textcolor{blue}{\textbf{0.0184}} & 0.0188 & \textcolor{red}{\textbf{0.0131}} & 0.0288 & 0.0683 & 0.0532 & 0.0626 & 0.0622 \\
 & $AUC_{OD}\uparrow$ & 1.0844 & 1.0565 & 1.2377 & 1.3562 & \textcolor{darkgreen}{\textbf{1.4148}} & 1.2762 & 1.2790 & 1.2974 & 1.2276 & \textcolor{red}{\textbf{1.4583}} & 1.3419 & \textcolor{blue}{\textbf{1.4034}} & 1.3259 \\
 & $AUC_{SNPR}\uparrow$ & 2.5734 & 2.3408 & 11.1531 & 14.2112 & \textcolor{red}{\textbf{29.1399}} & \textcolor{blue}{\textbf{16.6105}} & 16.4340 & \textcolor{darkgreen}{\textbf{24.4643}} & 9.0125 & 7.7506 & 7.5207 & 7.5270 & 6.3386 \\
\midrule
\multirow[c]{5}{*}{\shortstack[c]{HAD100-95\\\cite{li2023you}}} & $AUC_{(D,F)}\uparrow$ & 0.9806 & 0.8717 & 0.9620 & 0.9842 & 0.9917 & 0.9885 & \textcolor{blue}{\textbf{0.9920}} & \textcolor{darkgreen}{\textbf{0.9940}} & 0.9814 & 0.9919 & 0.9841 & 0.9762 & \textcolor{red}{\textbf{0.9944}} \\
 & $AUC_{(D,\tau)}\uparrow$ & 0.4106 & 0.1898 & 0.2016 & 0.3969 & 0.2202 & 0.3076 & 0.3824 & 0.3466 & 0.2608 & \textcolor{red}{\textbf{0.5199}} & 0.2458 & \textcolor{blue}{\textbf{0.4489}} & \textcolor{darkgreen}{\textbf{0.4735}} \\
 & $AUC_{(F,\tau)}\downarrow$ & 0.0930 & 0.0494 & 0.0182 & 0.0670 & \textcolor{darkgreen}{\textbf{0.0104}} & \textcolor{blue}{\textbf{0.0114}} & 0.0116 & \textcolor{red}{\textbf{0.0095}} & 0.0343 & 0.1055 & 0.0172 & 0.0826 & 0.0430 \\
 & $AUC_{OD}\uparrow$ & 1.2982 & 1.0121 & 1.1454 & 1.3141 & 1.2015 & 1.2847 & \textcolor{blue}{\textbf{1.3628}} & 1.3311 & 1.2078 & \textcolor{darkgreen}{\textbf{1.4062}} & 1.2126 & 1.3425 & \textcolor{red}{\textbf{1.4249}} \\
 & $AUC_{SNPR}\uparrow$ & 4.4139 & 3.8419 & 11.0683 & 5.9231 & 21.0954 & \textcolor{blue}{\textbf{27.0507}} & \textcolor{darkgreen}{\textbf{33.0728}} & \textcolor{red}{\textbf{36.6362}} & 7.5951 & 4.9267 & 14.2522 & 5.4326 & 11.0109 \\
\midrule
\multirow[c]{5}{*}{\shortstack[c]{Hyperion\\\cite{Hyperion}}} & $AUC_{(D,F)}\uparrow$ & 0.9829 & 0.9743 & 0.9849 & 0.9823 & 0.9958 & 0.9896 & 0.9877 & 0.9913 & \textcolor{blue}{\textbf{0.9987}} & 0.9985 & 0.9971 & \textcolor{red}{\textbf{0.9990}} & \textcolor{darkgreen}{\textbf{0.9988}} \\
 & $AUC_{(D,\tau)}\uparrow$ & 0.2251 & 0.4298 & 0.3624 & 0.4113 & 0.2140 & 0.3910 & 0.3793 & 0.3756 & 0.2453 & \textcolor{darkgreen}{\textbf{0.5534}} & 0.4658 & \textcolor{blue}{\textbf{0.4773}} & \textcolor{red}{\textbf{0.7706}} \\
 & $AUC_{(F,\tau)}\downarrow$ & 0.0434 & 0.0847 & 0.0200 & 0.0277 & \textcolor{red}{\textbf{0.0051}} & 0.0155 & 0.0134 & \textcolor{blue}{\textbf{0.0123}} & \textcolor{darkgreen}{\textbf{0.0107}} & 0.0809 & 0.0480 & 0.0574 & 0.0713 \\
 & $AUC_{OD}\uparrow$ & 1.1646 & 1.3193 & 1.3273 & 1.3658 & 1.2047 & 1.3651 & 1.3536 & 1.3546 & 1.2334 & \textcolor{darkgreen}{\textbf{1.4710}} & 1.4149 & \textcolor{blue}{\textbf{1.4189}} & \textcolor{red}{\textbf{1.6981}} \\
 & $AUC_{SNPR}\uparrow$ & 5.1840 & 5.0718 & 18.0884 & 14.8318 & \textcolor{red}{\textbf{42.1022}} & 25.2401 & \textcolor{blue}{\textbf{28.3871}} & \textcolor{darkgreen}{\textbf{30.5968}} & 23.0159 & 6.8422 & 9.7023 & 8.3211 & 10.8141 \\
\midrule
\multirow[c]{5}{*}{\shortstack[c]{Segundo\\\cite{Segundo}}} & $AUC_{(D,F)}\uparrow$ & 0.9841 & 0.8093 & 0.9545 & 0.9574 & 0.9873 & \textcolor{darkgreen}{\textbf{0.9920}} & 0.9896 & 0.9419 & 0.9872 & \textcolor{blue}{\textbf{0.9908}} & 0.9769 & 0.9708 & \textcolor{red}{\textbf{0.9943}} \\
 & $AUC_{(D,\tau)}\uparrow$ & 0.0119 & 0.0137 & 0.0728 & 0.0457 & 0.0179 & 0.0589 & 0.0499 & 0.0220 & 0.0192 & \textcolor{red}{\textbf{0.9498}} & \textcolor{blue}{\textbf{0.1082}} & 0.0522 & \textcolor{darkgreen}{\textbf{0.4719}} \\
 & $AUC_{(F,\tau)}\downarrow$ & 0.0017 & 0.0018 & 0.0014 & 0.0028 & \textcolor{blue}{\textbf{0.0005}} & \textcolor{blue}{\textbf{0.0005}} & \textcolor{darkgreen}{\textbf{0.0004}} & \textcolor{red}{\textbf{0.0003}} & 0.0016 & 0.7432 & 0.0182 & 0.0027 & \textcolor{blue}{\textbf{0.0005}} \\
 & $AUC_{OD}\uparrow$ & 0.9943 & 0.8212 & 1.0260 & 1.0002 & 1.0047 & 1.0505 & 1.0392 & 0.9636 & 1.0049 & \textcolor{darkgreen}{\textbf{1.1974}} & \textcolor{blue}{\textbf{1.0669}} & 1.0203 & \textcolor{red}{\textbf{1.4657}} \\
 & $AUC_{SNPR}\uparrow$ & 7.0471 & 7.7073 & 53.0914 & 16.2584 & 37.9194 & \textcolor{blue}{\textbf{128.6100}} & \textcolor{darkgreen}{\textbf{132.7147}} & 65.6002 & 12.2925 & 1.2780 & 5.9299 & 19.1691 & \textcolor{red}{\textbf{943.8000}} \\
\midrule
\multirow[c]{5}{*}{\shortstack[c]{UHAD-I\\\cite{liu2025adaptive}}} & $AUC_{(D,F)}\uparrow$ & 0.9237 & 0.8674 & 0.9163 & 0.8575 & 0.7457 & 0.9014 & 0.8863 & 0.8771 & \textcolor{darkgreen}{\textbf{0.9510}} & 0.9314 & \textcolor{blue}{\textbf{0.9387}} & 0.8767 & \textcolor{red}{\textbf{0.9660}} \\
 & $AUC_{(D,\tau)}\uparrow$ & \textcolor{blue}{\textbf{0.2681}} & 0.0448 & 0.1122 & 0.0059 & 0.2662 & 0.0828 & 0.0802 & 0.0524 & 0.1101 & \textcolor{red}{\textbf{0.9662}} & 0.2586 & 0.0764 & \textcolor{darkgreen}{\textbf{0.7191}} \\
 & $AUC_{(F,\tau)}\downarrow$ & 0.0366 & 0.0151 & 0.0074 & 0.0294 & 0.0292 & \textcolor{darkgreen}{\textbf{0.0060}} & \textcolor{blue}{\textbf{0.0065}} & \textcolor{red}{\textbf{0.0049}} & 0.0066 & 0.8957 & 0.0414 & 0.0340 & 0.0894 \\
 & $AUC_{OD}\uparrow$ & \textcolor{blue}{\textbf{1.1552}} & 0.8970 & 1.0211 & 0.8339 & 0.9827 & 0.9781 & 0.9601 & 0.9246 & 1.0545 & 1.0020 & \textcolor{darkgreen}{\textbf{1.1560}} & 0.9191 & \textcolor{red}{\textbf{1.5957}} \\
 & $AUC_{SNPR}\uparrow$ & 7.3251 & 2.9635 & \textcolor{darkgreen}{\textbf{15.1561}} & 0.1996 & 9.1276 & \textcolor{blue}{\textbf{13.7503}} & 12.3489 & 10.6614 & \textcolor{red}{\textbf{16.7194}} & 1.0788 & 6.2532 & 2.2474 & 8.0461 \\
\midrule
\multirow[c]{5}{*}{Average} & $AUC_{(D,F)}\uparrow$ & 0.9560 & 0.8911 & 0.9062 & 0.9572 & 0.9513 & 0.9711 & 0.9577 & 0.9443 & \textcolor{darkgreen}{\textbf{0.9757}} & 0.9380 & \textcolor{blue}{\textbf{0.9745}} & 0.9666 & \textcolor{red}{\textbf{0.9915}} \\
 & $AUC_{(D,\tau)}\uparrow$ & 0.2023 & 0.1810 & 0.1767 & 0.2126 & 0.2360 & 0.2175 & 0.2090 & 0.2087 & 0.1737 & \textcolor{red}{\textbf{0.7995}} & \textcolor{blue}{\textbf{0.2975}} & 0.2795 & \textcolor{darkgreen}{\textbf{0.6408}} \\
 & $AUC_{(F,\tau)}\downarrow$ & 0.0487 & 0.0684 & 0.0144 & 0.0243 & \textcolor{blue}{\textbf{0.0109}} & \textcolor{red}{\textbf{0.0106}} & \textcolor{darkgreen}{\textbf{0.0109}} & 0.0121 & 0.0198 & 0.5554 & 0.0438 & 0.0434 & 0.0576 \\
 & $AUC_{OD}\uparrow$ & 1.1095 & 1.0036 & 1.0684 & 1.0205 & 1.1763 & 1.1780 & 1.1558 & 1.1409 & 1.1296 & 1.1822 & \textcolor{darkgreen}{\textbf{1.2281}} & \textcolor{blue}{\textbf{1.2027}} & \textcolor{red}{\textbf{1.5748}} \\
 & $AUC_{SNPR}\uparrow$ & 5.3576 & 7.2775 & 25.9108 & 11.3387 & 27.3200 & \textcolor{darkgreen}{\textbf{77.3772}} & 52.1809 & \textcolor{blue}{\textbf{68.1779}} & 11.3671 & 3.1626 & 8.2347 & 8.5812 & \textcolor{red}{\textbf{127.1535}} \\
\bottomrule
\end{tabular}
}
\vspace*{-0.8\baselineskip}
\end{table*}

\subsubsection{Comparisons of Detection Maps}

\begin{figure*}[t]
    \centering
    \includegraphics[width=\textwidth]{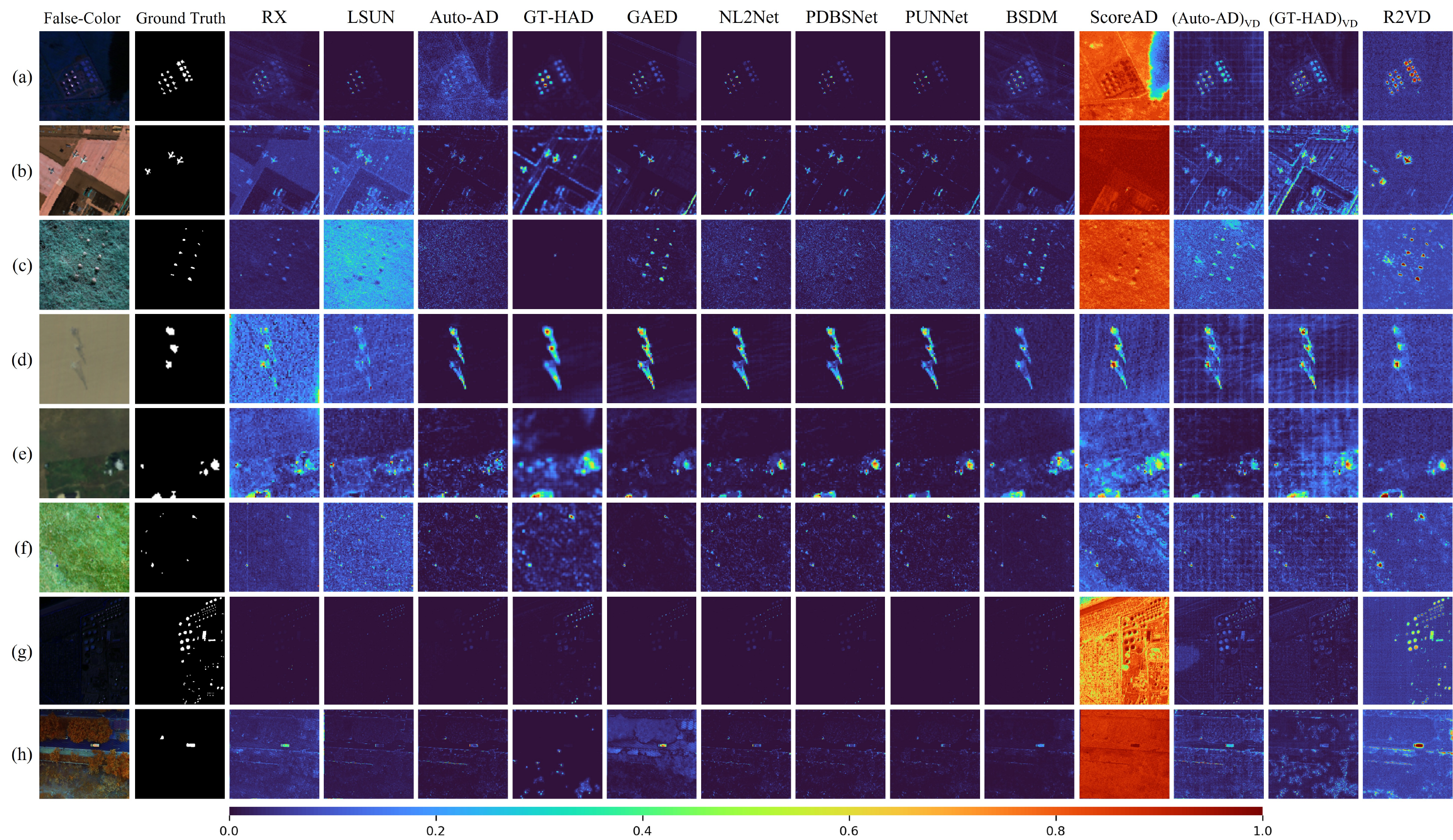}
    \vspace*{-2em}
    \caption{Visual comparison of anomaly detection maps on eight hyperspectral datasets. (a) Urban-2. (b) Aviris-2. (c) Cri. (d) HAD100-40. (e) HAD100-95. (f) Hyperion. (g) Segundo. (h) UHAD-I. R2VD distinctly highlights anomalies while facilitating the preservation of sub-pixel targets and topological structures.}
    \vspace*{-0.8\baselineskip}
    \label{fig:visual_comparison}
\end{figure*}

Fig. \ref{fig:visual_comparison} visually compares anomaly detection maps across eight hyperspectral datasets. The proposed R2VD demonstrates superior anomaly-background separation, maintaining a uniform low-response background while distinctly highlighting anomalies. Conversely, baselines like GT-HAD yield fragmented maps with inconsistent background suppression (e.g., Aviris-2). Consequently, in datasets like Cri, Hyperion, and UHAD-I, R2VD achieves stable target highlighting with reduced false alarms and missed detections.

Furthermore, R2VD exhibits high sensitivity to clustered anomalies often overlooked by conventional algorithms, successfully identifying them in Urban-2 and Segundo. For densely distributed anomalies in HAD100-95, R2VD precisely isolates individual targets, unlike GT-HAD and ScoreAD which tend to merge them. Since ScoreAD also uses score-based vector information, this demonstrates that vector inference within a purified residual space effectively mitigates the over-smoothing of local spatial features.

R2VD also demonstrates robustness against environmental interferences. In HAD100-40, it suppresses shadows to minimize misclassifications. Although the Aviris-2 map shows minor false alarms and partial missed detections on airplane wings, R2VD's overall background uniformity remains highly competitive. This trade-off validates the PSF's efficacy in strictly restricting background contamination within challenging scenes.

\subsubsection{Comparisons of Separability Maps}

\begin{figure*}[t]
    \centering
    \includegraphics[width=\textwidth]{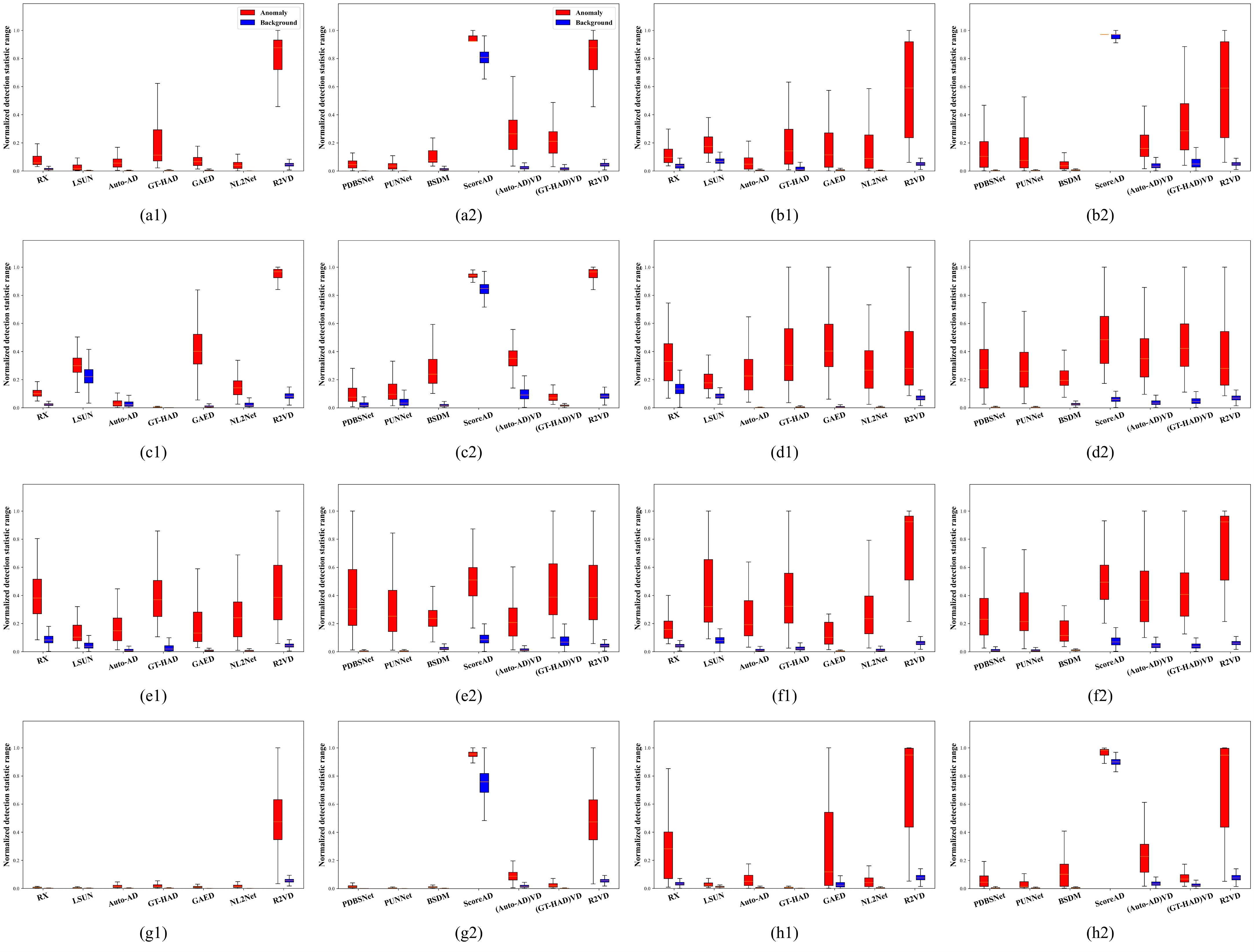}
    \vspace*{-2em}
    \caption{Statistical separability of anomaly (red) and background (blue) pixels across eight datasets, evaluated by normalized detection statistics. Boxes and whiskers indicate the interquartile range and statistical extent across thirteen methods, including the proposed R2VD, with the methods displayed in two subpanels for each dataset. (a1)--(a2) Urban-2. (b1)--(b2) Aviris-2. (c1)--(c2) Cri. (d1)--(d2) HAD100-40. (e1)--(e2) HAD100-95. (f1)--(f2) Hyperion. (g1)--(g2) Segundo. (h1)--(h2) UHAD-U-I. Results show R2VD yields competitive anomaly-background separation.}
    \vspace*{-0.5\baselineskip}
    \label{fig:visual_boxplot}
\end{figure*}

To intuitively evaluate the statistical separability between backgrounds and anomalies, Fig. \ref{fig:visual_boxplot} visualizes the distribution of the final detection scores via boxplots. For each boxplot, the lower and upper box edges denote the first and third quartiles, respectively, and the box height represents the interquartile range (IQR). The central line denotes the median, while the whiskers extend to the most extreme values within $1.5\times\mathrm{IQR}$ below the first quartile and above the third quartile. Values beyond these limits are treated as outliers and are omitted for visual clarity. In a highly discriminative model, the background distribution should be tightly compressed near zero to reflect effective suppression, while the anomaly distribution should span a higher score interval to indicate robust target highlighting. Across all eight evaluated datasets, the proposed R2VD framework consistently exhibits this optimal behavior. Taking the complex Segundo dataset as a prime example, the competing baseline methods suffer from severe confirmation bias where both their background and anomaly score distributions collapse into indistinguishable flat boxes near the zero baseline. In contrast, our R2VD resolves this bottleneck by maintaining a rigorously suppressed compact blue box alongside an activated elongated red box at a higher position. This clear separation occurs because the vector dynamics process in the purified residual space highlights targets through the deterministic vector addition effect. Concurrently, The PSF reduces the tendency of the background score model to absorb anomalous characteristics.

\subsection{Statistical Reliability Evaluation}\label{IV-C}

To fully address the potential stochastic variance introduced by the $K$ independent microscopic perturbations in the VDI stage, we conduct a rigorous statistical reliability evaluation. Specifically, the proposed R2VD framework is executed across five entirely distinct random seeds (i.e., 42, 3407, 114514, 2026, and 8888) on three representative datasets: Aviris-2, HAD100-95, and Segundo, which encompass diverse environmental scenarios and target characteristics.

\begin{table}[t]
\centering
\caption{Statistical Reliability Analysis of the R2VD Over Five Random Seeds.}
\vspace*{-0.4\baselineskip}
\label{tab:reliability}
\footnotesize 
\setlength{\tabcolsep}{2pt} 
\begin{tabular*}{\columnwidth}{@{\extracolsep{\fill}}lccc}
\toprule
Metric & Aviris-2 & HAD100-95 & Segundo \\
\midrule
$AUC_{(D,F)}$ & 0.9865 $\pm$ 0.0012 & 0.9944 $\pm$ 0.0010 & 0.9932 $\pm$ 0.0008 \\
$AUC_{(D,\tau)}$  & 0.5507 $\pm$ 0.0125 & 0.5015 $\pm$ 0.0173 & 0.4618 $\pm$ 0.0130 \\
$AUC_{(F,\tau)}$  & 0.0500 $\pm$ 0.0035 & 0.0446 $\pm$ 0.0019 & 0.0008 $\pm$ 0.0002 \\
$AUC_{OD}$          & 1.4872 $\pm$ 0.0142 & 1.4513 $\pm$ 0.0163 & 1.4543 $\pm$ 0.0133 \\
$AUC_{SNPR}$        & 11.0700 $\pm$ 0.8100    & 11.2500 $\pm$ 0.4800    & 638.6604 $\pm$ 180.3332 \\
\bottomrule
\end{tabular*}
\vspace*{-1\baselineskip}
\end{table}

The quantitative results, including the mean and standard deviation of the five foundational 3D ROC metrics, are summarized in Table. \ref{tab:reliability}. Crucially, the primary separability metric $AUC_{(D,F)}$ exhibits stability, yielding standard deviations as low as $0.0012$, $0.0010$, and $0.0008$ for the three datasets, respectively. Meanwhile, the threshold-dependent metrics ($AUC_{(D,\tau)}$ and $AUC_{(F,\tau)}$) also maintain remarkably narrow fluctuation intervals. This tight error margin provides solid empirical evidence that the superior performance of R2VD is reproducible and inherently driven by the robust score-field vector dynamics rather than random initialization. It statistically substantiates that within the purified residual space, the random omnidirectional perturbations consistently trigger robust vector cancellation for background clutter and deterministic alignment for anomalous targets, thereby ensuring the model's reliability in practical HAD.

To examine the perturbation-based vector aggregation mechanism, we fix the PPE/GMP outputs and trained DiT for each dataset, vary only the VDI noise over the same five seeds, and compare the low-count setting $K=5$ with the full setting $K=50$. Here, $K=5$ is the smallest non-degenerate setting. For the pre-normalization directional response, the Background Stability Variance (BSV) is defined as
\vspace*{-0.2\baselineskip}
\begin{equation}
\mathrm{BSV}(K)=\frac{1}{N_0}\sum_{i:y_i=0}
\operatorname{Var}_{m=1}^{M}\left(
\left\|\frac{1}{K}\sum_{k=1}^{K}\mathbf{u}_i^{(m,k)}\right\|_2
\right),
\label{eq:bsv}
\end{equation}
where $N_0$ is the number of background pixels and $M=5$ is the number of noise seeds. In addition, $\mathrm{FAR}_{95}$ denotes the false-alarm rate at $P_D=0.95$. As summarized in Table~\ref{tab:vdi_stability}, increasing $K$ from $5$ to $50$ consistently reduces BSV by $83.85\%$--$87.24\%$ and $\mathrm{FAR}_{95}$ by $85.50\%$--$95.12\%$, while improving $AUC_{(D,F)}$ by $0.0325$--$0.0848$. These controlled results verify that multi-perturbation VDI stabilizes background responses through directional cancellation while preserving coherent directional accumulation for anomalous residuals.

\begin{table}[t]
\caption{Perturbation stability comparison of VDI with $K=5$ and $K=50$ over five noise seeds.}
\label{tab:vdi_stability}
\centering
\footnotesize
\setlength{\tabcolsep}{2.8pt}
\renewcommand{\arraystretch}{1.25}
\begin{tabular}{llccc}
\hline
Dataset &
$K$ &
$AUC_{(D,F)}\uparrow$ &
BSV $\downarrow$ &
$\mathrm{FAR}_{95}\downarrow$ \\
\hline
\multirow[c]{2}{*}{Aviris-2}
& 5
& 0.9017 $\pm$ 0.0108
& 7.72${\times}10^{-4}$
& 0.5818 $\pm$ 0.0851 \\
& 50
& 0.9865 $\pm$ 0.0012
& 1.15${\times}10^{-4}$
& 0.0844 $\pm$ 0.0134 \\
\hline
\multirow[c]{2}{*}{HAD100-95}
& 5
& 0.9619 $\pm$ 0.0081
& 7.55${\times}10^{-4}$
& 0.2587 $\pm$ 0.0832 \\
& 50
& 0.9944 $\pm$ 0.0010
& 9.63${\times}10^{-5}$
& 0.0126 $\pm$ 0.0022 \\
\hline
\multirow[c]{2}{*}{Segundo}
& 5
& 0.9440 $\pm$ 0.0024
& 7.55${\times}10^{-4}$
& 0.4482 $\pm$ 0.0401 \\
& 50
& 0.9932 $\pm$ 0.0008
& 1.22${\times}10^{-4}$
& 0.0459 $\pm$ 0.0106 \\
\hline
\end{tabular}
\vspace*{-0.8\baselineskip}
\end{table}

\subsection{Ablation Studies and Analysis}\label{IV-D}

\begin{figure}[t]
    \centering
    \includegraphics[width=0.48\textwidth]{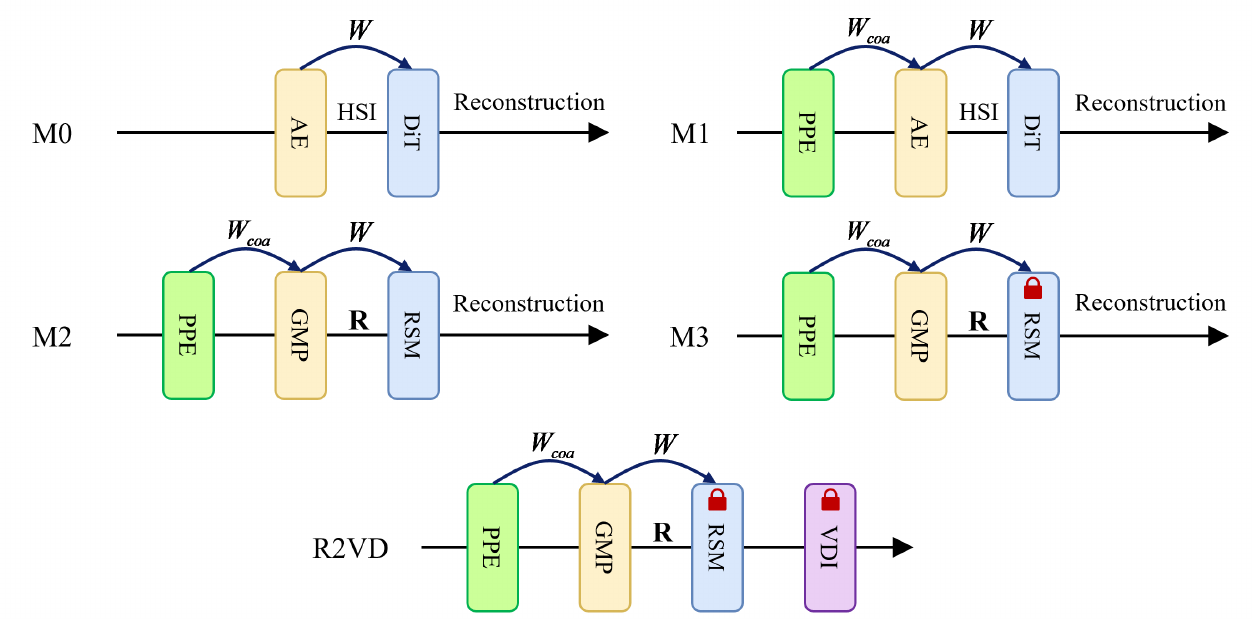}
    \caption{Structural illustration of the progressively constructed ablation variants.}
    \label{fig:ablation_flowchart}
    \vspace*{-0.8\baselineskip}
\end{figure}

In this section, we evaluate the contributions of the PPE, GMP, PSF, and VDI stages to the proposed method through a progressive ablation study. These incrementally constructed variants are structuralized in Fig. \ref{fig:ablation_flowchart}. The baseline model M0 utilizes an unguided autoencoder and a standard DiT architecture for scalar reconstruction directly on the raw hyperspectral image. Variant M1 incorporates the initial weight map $\mathbf{W}_{coa}$ from the PPE to guide the autoencoder. Variant M2 introduces the GMP to shift the diffusion training into the purified residual space $\mathbf{R}$, while variant M3 activates the hard constant penalty within the PSF to restrict cross-spectral information leakage during score field modeling. The final R2VD configuration completes the framework by substituting the scalar reconstruction with the VDI stage. All variants are evaluated under identical configurations to ensure an objective evaluation.

To complement $AUC_{(D,F)}$, we further report $AUC_{SNPR}$ and a mean anomaly--background Separability Score (SS) for each ablation variant. $AUC_{SNPR}$ quantifies the target-response advantage relative to background false-alarm responses across decision thresholds, whereas SS directly measures the mean separation in the final normalized detection map. Specifically, SS is defined as
\begin{equation}
\mathrm{SS}=
\frac{1}{N_1}\sum_{i=1}^{N}y_i a_i
-
\frac{1}{N_0}\sum_{i=1}^{N}(1-y_i)a_i,
\label{ablation_ss}
\end{equation}
where $a_i\in[0,1]$ is the normalized detection score, and $N_1$ and $N_0$ denote the numbers of anomaly and background pixels, respectively. A larger SS indicates a wider mean separation between anomaly and background responses.

\begin{figure}[t]
    \centering
    \includegraphics[width=0.48\textwidth]{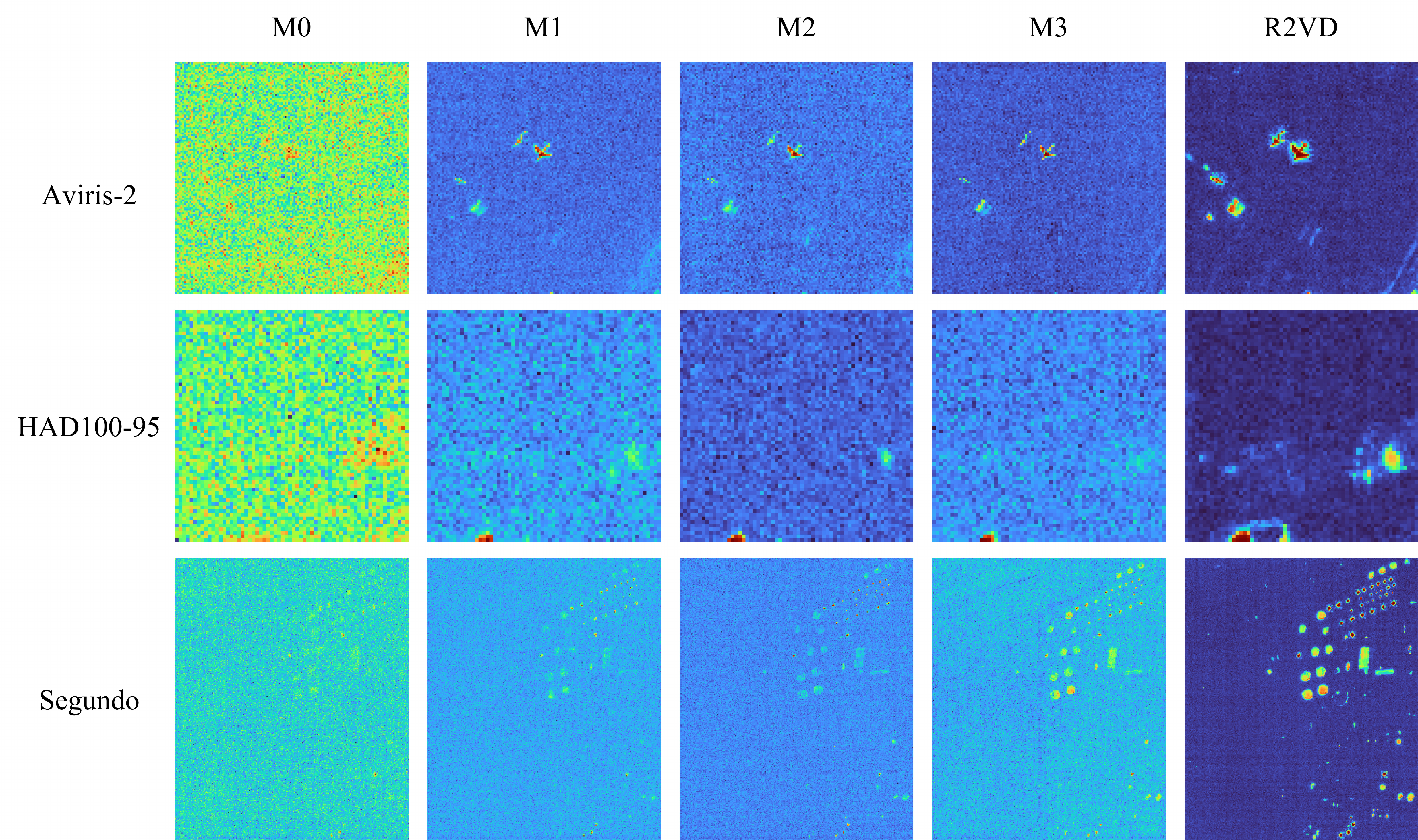}
    \caption{Visual evolution of anomaly detection maps across progressive ablation variants. Results confirm that each component mitigates background interference and enhances target-background contrast.}
    \vspace*{-0.8\baselineskip}
    \label{fig:ablation_vision}
\end{figure}

Table \ref{tab:ablation} and Fig. \ref{fig:ablation_vision} present the quantitative and visual results of the progressive ablation study. In addition to the primary $AUC_{(D,F)}$ metric, the added $AUC_{SNPR}$ and SS metrics quantify background suppression and anomaly--background score separation, respectively. Across Aviris-2, HAD100-95, and Segundo, all three metrics improve consistently from M0 to the complete R2VD configuration.

The baseline M0 yields the lowest $AUC_{(D,F)}$ values, together with low $AUC_{SNPR}$ values of $1.226$, $1.268$, and $1.179$ and SS values of $0.103$, $0.110$, and $0.053$ on the three datasets, respectively. Incorporating PPE in M1 produces initial improvements by reducing the effect of coarse anomaly contamination. GMP further improves both supplementary metrics in M2 by refining the background manifold and suppressing confirmation bias. With PSF enabled, M3 produces a further increase in $AUC_{SNPR}$ and SS, indicating more effective background suppression and a wider score gap between anomalies and backgrounds.

Finally, the complete R2VD configuration achieves the best results on all three datasets. Its $AUC_{SNPR}$ reaches $10.196$, $11.011$, and $943.800$, while SS reaches $0.513$, $0.399$, and $0.430$ on Aviris-2, HAD100-95, and Segundo, respectively. More specifically, the transition from M1 to M2 examines the contribution of GMP-based residual purification, since M2 shifts score modeling from the raw HSI to the purified residual space while retaining scalar scoring. This transition consistently improves $AUC_{(D,F)}$, $AUC_{SNPR}$, and SS on all three datasets. The transition from M3 to R2VD then evaluates the effect of replacing the scalar decision with VDI under the same purified-residual and PSF-constrained score-modeling setting. The resulting gains across all three metrics indicate that GMP-based residual purification and VDI provide complementary contributions to anomaly-background separation.

\begin{table}[!t]
\caption{Quantitative ablation study of the core components in the proposed R2VD framework.}
\label{tab:ablation}
\centering
\footnotesize
\setlength{\tabcolsep}{2.8pt}
\renewcommand{\arraystretch}{1.08}
\begin{tabular}{llccc}
\hline
Dataset & Model &
$AUC_{(D,F)}\uparrow$ &
$AUC_{SNPR}\uparrow$ &
SS$\uparrow$ \\
\hline
\multirow[c]{5}{*}{Aviris-2}
& M0   & 0.7108 & 1.226 & 0.1035 \\
& M1   & 0.7851 & 2.786 & 0.1311 \\
& M2   & 0.8254 & 3.196 & 0.1712 \\
& M3   & 0.8930 & 5.626 & 0.2671 \\
& R2VD & \textbf{0.9878} & \textbf{10.196} & \textbf{0.5127} \\
\hline
\multirow[c]{5}{*}{HAD100-95}
& M0   & 0.7233 & 1.268 & 0.1096 \\
& M1   & 0.8121 & 2.507 & 0.1430 \\
& M2   & 0.8367 & 3.027 & 0.1697 \\
& M3   & 0.8952 & 3.733 & 0.2232 \\
& R2VD & \textbf{0.9944} & \textbf{11.011} & \textbf{0.3988} \\
\hline
\multirow[c]{5}{*}{Segundo}
& M0   & 0.6766 & 1.179 & 0.0531 \\
& M1   & 0.7640 & 1.326 & 0.0684 \\
& M2   & 0.8149 & 2.524 & 0.0888 \\
& M3   & 0.8646 & 7.655 & 0.1539 \\
& R2VD & \textbf{0.9943} & \textbf{943.800} & \textbf{0.4295} \\
\hline
\end{tabular}
\vspace*{-0.8\baselineskip}
\end{table}

\subsection{Parameter Sensitivity Analysis}

\begin{figure}[t]
    \centering
    \includegraphics[width=0.48\textwidth]{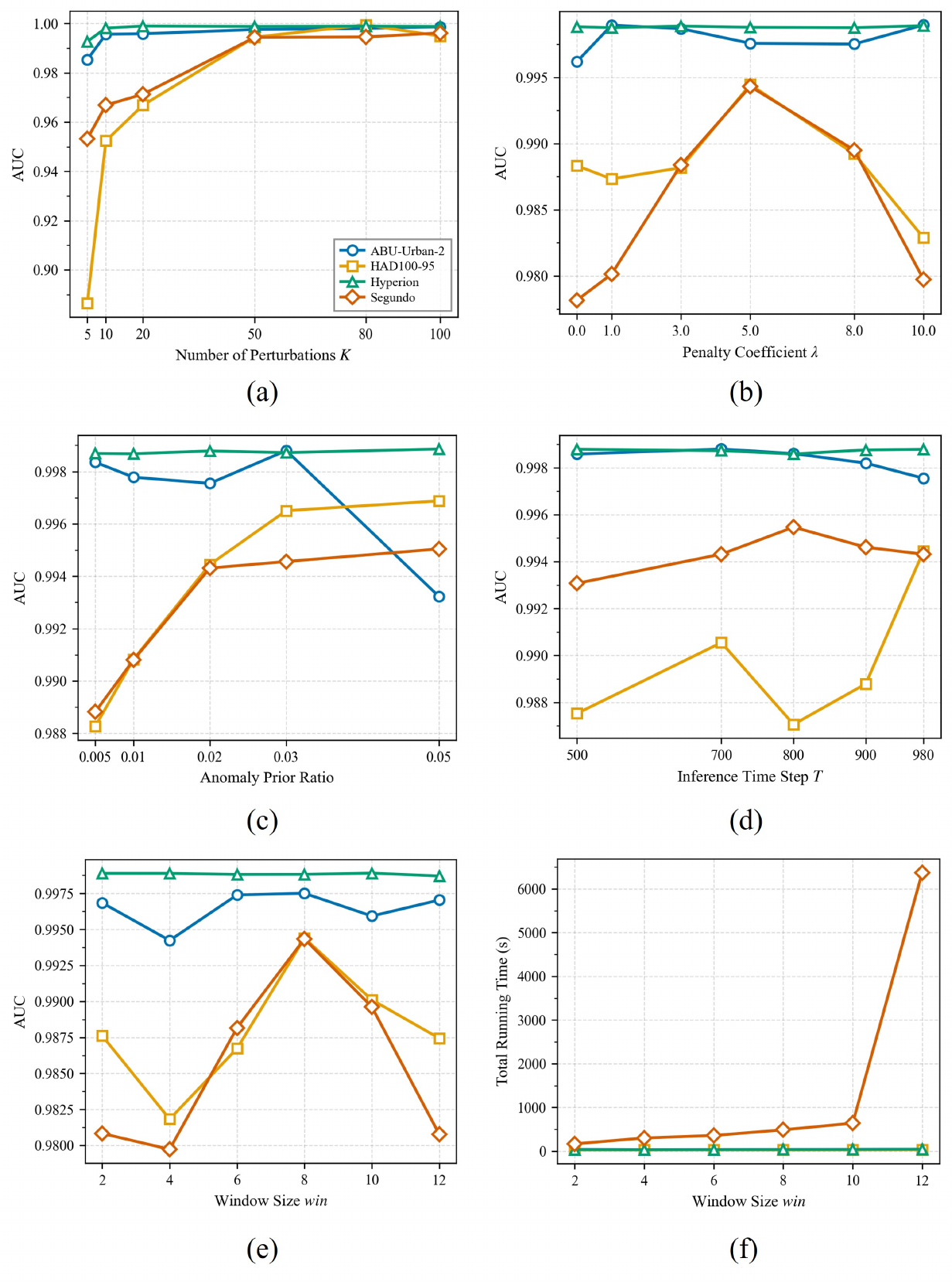} 
    \vspace*{-0.5\baselineskip}
    \caption{Hyperparameter sensitivity analysis across four HSI datasets. (a) Micro-perturbation count $K$. (b) PSF penalty constant $\lambda$. (c) Global anomaly prior ratio $\eta$. (d) Inference time step $t_{\text{inf}}$. (e) Window size $win$. (f) Corresponding running time under different window sizes. The shared legend is in (a). Stable trajectories demonstrate the framework's robustness and adaptive error correction.}
    \vspace*{-0.8\baselineskip}
    \label{fig:hyperparameter_sensitivity}
\end{figure}

To validate the robustness of the R2VD framework and examine the impact of its physical mechanisms on detection performance, we perform a sensitivity analysis on five core hyperparameters. The performance variations across the Urban-2, HAD100-95, Hyperion, and Segundo datasets are illustrated in Fig. \ref{fig:hyperparameter_sensitivity}.

The perturbation count $K$ in the VDI stage determines the sample size for revealing the score field direction. Fig. \ref{fig:hyperparameter_sensitivity}(a) shows $AUC_{(D,F)}$ values initially increasing before saturating. On HAD100-95, this metric rises from 0.8864 to 0.9944 as $K$ approaches 50. This confirms that larger $K$ values stabilize score vectors, enabling destructive background interference and constructive anomaly interference. We recommend $K=50$ to optimally balance accuracy and computational cost.

The physical penalty constant $\lambda$ in the PSF regulates feature leakage between heterogeneous signatures. Fig.\ref{fig:hyperparameter_sensitivity}(b) further includes $\lambda=0$, at which the PSF penalty in Eq.~\eqref{attn} is removed and the RSM module uses standard window self-attention within the otherwise unchanged R2VD backbone. Across the four representative datasets, introducing a nonzero PSF penalty consistently yields the best result at a positive $\lambda$. Although the optimal value is scene dependent, $\lambda=5$ achieves the highest average $AUC_{(D,F)}$ of $0.9963$, compared with $0.9904$ for $\lambda=0$. In particular, the improvements on HAD100-95 and Segundo are $0.0061$ and $0.0161$, respectively. These results indicate that the PSF effectively suppresses cross-spectral feature leakage, whereas excessively strong penalties may restrict useful background-context interactions. Therefore, $\lambda=5$ is adopted as a robust unified setting without scene-specific tuning. A moderate penalty mitigates cross-spectral feature leakage while preserving useful background-context interactions. Conversely, an overly strict penalty severs intrinsic background correlations, impairing global modeling. Thus, $\lambda=5.0$ serves as an effective empirical equilibrium across diverse scenes.

The anomaly prior ratio $\eta$ initializes the anomaly proportion in the PPE stage. Fig. \ref{fig:hyperparameter_sensitivity}(c) demonstrates stable, high-level performance across all datasets. This insensitivity highlights the GMP stage's adaptive error correction. Driven by dynamic weight map $\mathbf{W}$ updates, the model autonomously rectifies manifold boundaries through iterative purification, allowing R2VD to excel without precise scene-specific tuning of $\eta$.

The inference time step $t_{inf}$ dictates the noise scale in the frozen score field. Fig. \ref{fig:hyperparameter_sensitivity}(d) shows stability across a broad 500–980 interval. Even at extreme noise levels ($t_{inf}=980$), R2VD maintains performance and avoids the structural collapse typical of traditional diffusion models. This robustness confirms the PSF effectively prevents erroneous inter-manifold information exchange under deep perturbations.

The window size $win$ determines the spatial range of local attention in the RSM module. As shown in Fig.~\ref{fig:hyperparameter_sensitivity}(e), $win=8$ achieves the highest average $AUC_{(D,F)}$ of $0.9963$ and yields the best results on three datasets. Although Hyperion attains a marginally higher value at $win=10$, the difference from $win=8$ is only $0.0001$. Small windows provide insufficient contextual information, whereas overly large windows introduce heterogeneous pixels into the same attention region and can weaken anomaly-background discrimination. Fig.~\ref{fig:hyperparameter_sensitivity}(f) further shows that the running time generally increases with window size, consistent with the $\mathcal{O}(HW\cdot win^2)$ complexity of local attention and PSF distance computation. This effect is particularly evident on Segundo, where increasing $win$ from $8$ to $12$ raises the running time from $487.04$~s to $6371.70$~s without improving detection performance. Therefore, $win=8$ is adopted as a balanced setting for contextual modeling accuracy and computational efficiency.

\subsection{Complexity and Efficiency Analysis}\label{sec:con}

To provide a comprehensive efficiency evaluation, Table \ref{tab:complexity_comparison} reports the learnable parameters and end-to-end running time of all compared methods on the Urban-2 dataset. The reported time includes both training and testing. RX and LSUN are denoted by ``--'' because they have no learnable parameters.

Traditional detectors and lightweight reconstruction networks require relatively limited computation. RX, Auto-AD, and GT-HAD require 0.18 s, 4.60 s, and 6.80 s, respectively. The self-supervised reconstruction methods require more time, ranging from 36.13 s for GAED to 138.69 s for NL2Net, due to their contextual modeling and optimization procedures. The two plug-in variants, (Auto-AD)$_{VD}$ and (GT-HAD)$_{VD}$, require 20.00 s and 67.80 s, respectively, further demonstrating the compatibility of the proposed RSM and VDI stages with different reconstruction priors.

R2VD contains 1.27 M learnable parameters and requires 37.54 s for end-to-end processing. Although residual score modeling with the DiT introduces additional computation relative to lightweight autoencoder baselines, R2VD is more efficient than PUNNet, PDBSNet, NL2Net, and ScoreAD. In particular, R2VD avoids the iterative reverse diffusion sampling required by ScoreAD and thus substantially reduces its running time from 452.15 s to 37.54 s, while achieving state-of-the-art detection performance.

\begin{table}[t]
\centering
\caption{Complexity and end-to-end running time comparison on the Urban-2 dataset.}
\vspace*{-0.2\baselineskip}
\label{tab:complexity_comparison}
\footnotesize
\renewcommand{\arraystretch}{0.95}
\begin{tabular}{lcc}
\toprule
Method & Params (M) & Total Running Time (s) \\
\midrule
RX \cite{RX}                         & --   & 0.18 \\
LSUN \cite{LSUN}                     & --   & 7.04 \\
Auto-AD \cite{wang2021auto}          & 3.25 & 4.60 \\
GT-HAD \cite{lian2024gt}             & 0.26 & 6.80 \\
GAED \cite{gaed}                     & 0.05 & 36.13 \\
NL2Net \cite{NL2Net}                 & 0.77 & 138.69 \\
PDBSNet \cite{PDBSNet}               & 0.55 & 68.43 \\
PUNNet \cite{PUNNet}                 & 2.34 & 43.80 \\
BSDM \cite{ma2025bsdm}               & 1.16 & 18.16 \\
ScoreAD \cite{ScoreAD}               & 0.54 & 452.15 \\
(Auto-AD)$_{VD}$              & 4.52 & 20.00 \\
(GT-HAD)$_{VD}$               & 1.53 & 67.80 \\
\midrule
R2VD (Ours)                 & 1.27 & 37.54 \\
\bottomrule
\end{tabular}
\vspace*{-0.8\baselineskip}
\end{table}

\subsection{Intrinsic Mechanism of the PSF}

\begin{figure}[t]
    \centering
    \includegraphics[width=\linewidth]{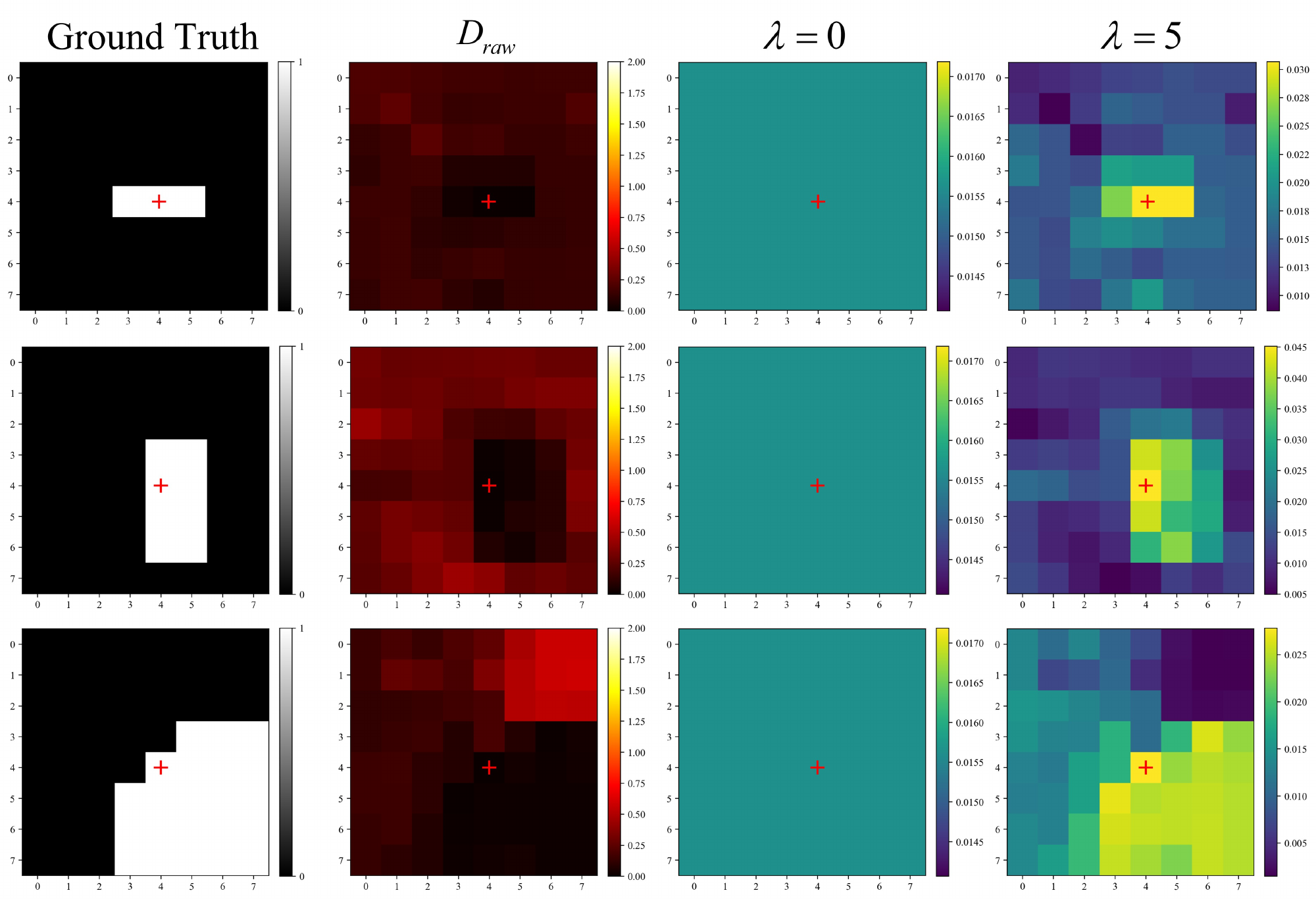} 
    \caption{Visual analysis of the PSF mechanism across varying anomaly scales. By incorporating the physical penalty ($\lambda=5$), the PSF contributes to the formation of highly localized spectral clusters.}
    \vspace*{-0.5\baselineskip}
    \label{fig:psf_analysis}
\end{figure}

To address the limitations of scalar metrics and prevent the loss of directional information during generative modeling, we investigate the intrinsic mechanism of the PSF. We analyze the micro-level attention distributions across tiny, medium, and large anomaly clusters, corresponding to the centers (r86, c263), (r86, c235), and (r92, c94) in the Segundo dataset, respectively. As illustrated in Fig. \ref{fig:psf_analysis}, darker regions in the spectral distance matrix $\mathbf{D}_{raw}$ represent closer spectral proximity, whereas brighter colors in the attention maps indicate higher attention weights. Across all three rows, $\mathbf{D}_{raw}$ reveals that the query anomaly (denoted by the red cross) maintains high spectral similarity with adjacent anomalous pixels (darker areas) but exhibits a certain degree of divergence from the surrounding background (brighter areas). When the PSF is deactivated ($\lambda=0$), attention weights diffuse uniformly across the spatial window, causing anomalous features to be averaged out by the background statistics. Conversely, upon integrating the physical penalty ($\lambda=5$), the attention mechanism is effectively regularized. Consequently, the query pixel assigns notably high attention (bright areas) exclusively to its spectrally similar anomalous neighbors while effectively suppressing interactions with the spectrally distant background (dark areas). This consistent behavior indicates that the PSF facilitates the formation of highly localized spectral clusters within the attention field, thereby mitigating feature leakage and establishing a reliable foundation for subsequent VDI.

\subsection{Preservation of Sub-pixel Anomalies}

\begin{figure}[t]
    \centering
    \includegraphics[width=\linewidth]{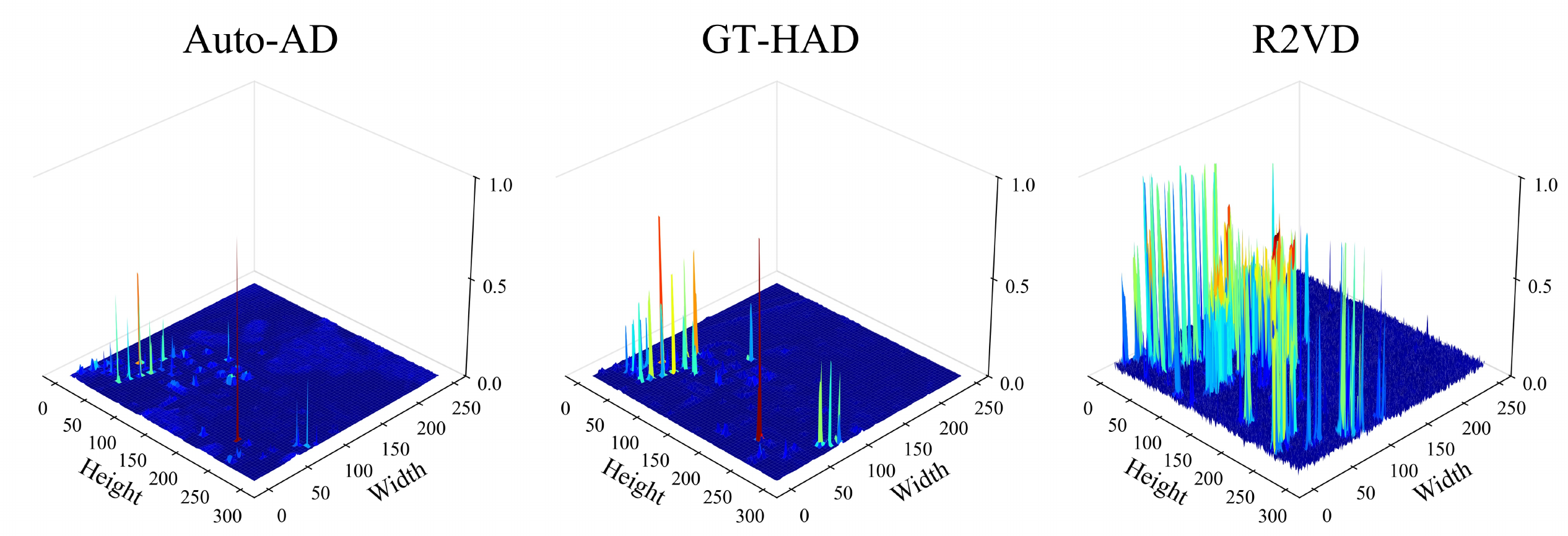} 
    \caption{Three-dimensional anomaly score response maps on the Segundo dataset. By avoiding spatial smoothing, R2VD mitigates the attenuation of miniature anomalies.}
    \vspace*{-0.8\baselineskip}
    \label{fig:3d_response}
\end{figure}

Addressing the second intrinsic defect identified in traditional paradigms, we evaluate the framework's capacity to preserve high-frequency sub-pixel topologies without relying on spatial downsampling. We visualize the three-dimensional anomaly score response maps on the Segundo dataset, as illustrated in Fig. \ref{fig:3d_response}. Traditional reconstruction-based autoencoders, such as Auto-AD and GT-HAD, are often susceptible to the aforementioned spatial smoothing effects induced by downsampling operations. Consequently, these methods frequently overlook numerous sub-pixel targets or yield heavily attenuated response amplitudes. Conversely, the proposed full-resolution architecture preserves these miniature anomalies more comprehensively, manifesting as a dense distribution of sharp and localized response peaks in the visualization. This comparison indicates that maintaining spatial fidelity during manifold purification effectively mitigates the vanishing of fragile topologies and facilitates more robust target-background decoupling.

\subsection{Confirmation Bias Analysis}

\begin{figure}[t]
    \centering
    \includegraphics[width=0.48\textwidth]{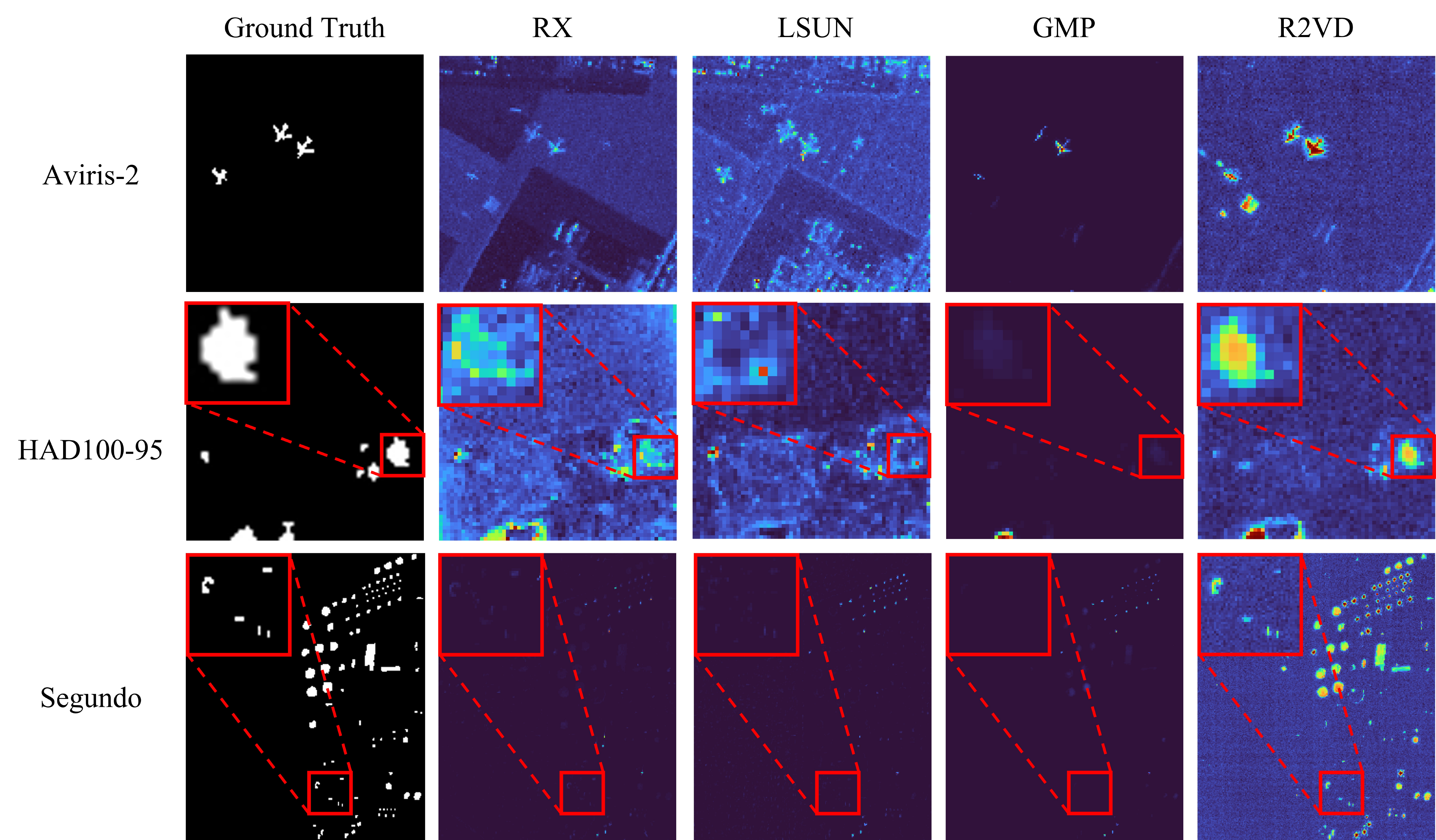}
    \caption{Visual comparison of detection maps across different stages. Red boxes and dashed lines indicate magnified regions. While the intermediate GMP stage effectively suppresses the prominent false alarms inherited from coarse detectors, it inevitably smooths out fragile target profiles. The complete R2VD framework successfully rectifies this structural attenuation, recovering sharp and high-contrast anomaly topologies.}
    \vspace*{-0.8\baselineskip}
    \label{fig:prior_comparison}
\end{figure}

\begin{figure}[t]
    \centering
    \includegraphics[width=0.48\textwidth]{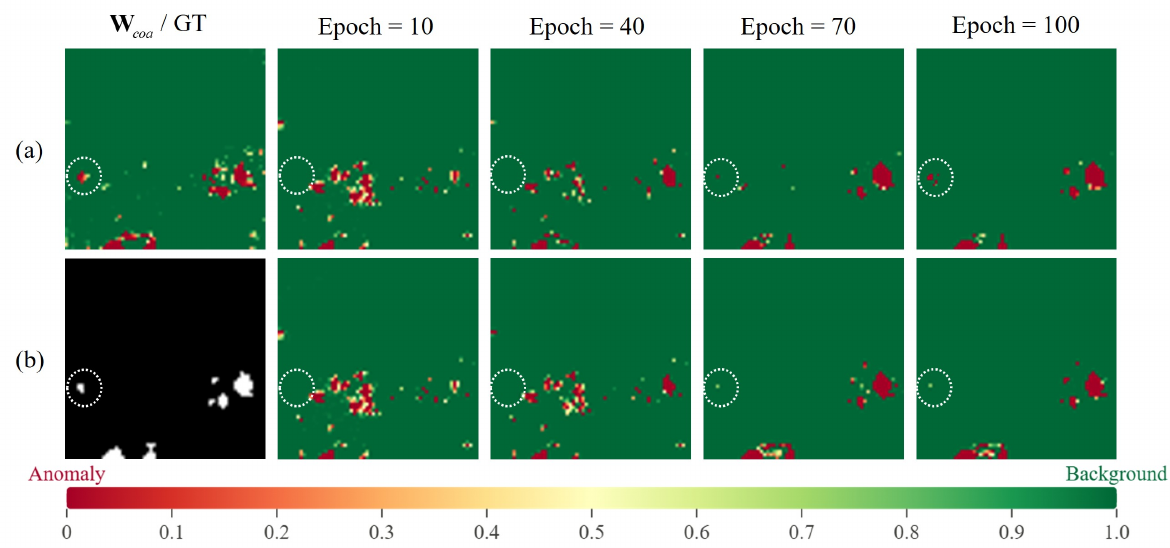}
    \caption{Visual comparison of weight map dynamics on the HAD100-95 dataset. Row (a) illustrates the purification process guided by the initial physical prior ($\mathbf{W}_{coa}$). Row (b) shows the unguided autoencoder baseline against the Ground Truth. The synergy between PPE and GMP effectively self-corrects cognitive biases, generating a more robust gradient-gating mask for subsequent score field modeling.}
    \vspace*{-0.8\baselineskip}
    \label{fig:bias}
\end{figure}

To evaluate robustness against the confirmation bias inherent in unsupervised training, Fig. \ref{fig:prior_comparison} and Fig. \ref{fig:bias} visualize the synergistic optimization and structural evolution across consecutive stages. As demonstrated by the intermediate GMP results in Fig. \ref{fig:prior_comparison}, although classical RX and LSUN algorithms provide imperfect or noisy coarse footprints, the guided autoencoder exhibits remarkable resilience to these initial misalignments, substantially suppressing prominent background false alarms on Aviris-2. However, due to the implicit low-pass filtering effect of autoencoder architectures, the purified outputs of the GMP stage tend to smooth out subtle high-frequency features, causing fragile sub-pixel anomalies to severely attenuate or vanish on HAD100-95 and Segundo. Crucially, this structural limitation fully justifies the necessity of the subsequent generative modeling; by passing this purified space into the PSF-constrained DiT, the complete R2VD framework effectively captures high-dimensional vector dynamics to recover sharp target topologies and achieve optimal target-background contrast. This error-correction capability is explicitly governed by the interaction within the purification loop, as illustrated by the weight map dynamics in Fig. \ref{fig:bias}. Under explicit physical prior guidance, the dynamic weights initially suffering from alignment gaps progressively rectify target boundaries and eliminate background clutter by Epoch 100 (row a), whereas the unguided baseline undergoes severe target omission and anomaly memorization (row b). This mutual optimization guarantees that the purified residual space passed to the subsequent diffusion modeling maintains high-confidence background quality while isolating out-of-distribution anomaly states.

\section{Conclusion}\label{sec:con} 

In this paper, we proposed the R2VD framework to address the confirmation bias and sub-pixel anomaly vanishing prevalent in HAD. By extending the reconstruction-based approach to serve as a multi-stage decoupled pipeline, the framework utilizes reconstruction exclusively for background manifold purification, followed by high-dimensional vector dynamics inference. Initialized by physical priors from PPE, the GMP stage extracts a highly purified residual space while preserving fragile target topologies. Operating within this purified space, the combination of RSM constrained by the PSF and VDI effectively decouples anomalies from complex backgrounds by evaluating probability gradient directional stability instead of relying on ambiguous scalar residuals. Extensive experiments verified that our approach achieves superior background suppression and target highlighting, yielding a state-of-the-art primary separability metric near 0.99 across multiple datasets. 
Despite its strong performance, R2VD still has practical trade-offs, including the empirical setting of the initial anomaly ratio $\eta$ and higher computational cost from dual-stage DiT optimization. Its directional consistency may also weaken under extreme spectral overlap. Future work will develop adaptive prior estimation to automatically determine $\eta$ for fully unsupervised blind detection.

\bibliographystyle{IEEEtran}
\bibliography{ref}

\vfill

\end{document}